\documentclass[letterpaper,twocolumn,10pt]{article}

\PassOptionsToPackage{hyphens}{url}
\usepackage{usenix-2020-09}

\usepackage{tikz}
\usetikzlibrary{shapes, positioning, calc, patterns}
\usepackage{scalerel}

\usepackage{amsmath}
\usepackage{amssymb}
\usepackage{multirow}
\usepackage{caption}
\usepackage{subcaption}
\usepackage{booktabs}
\usepackage{enumitem}
\usepackage{algorithm}
\usepackage{algpseudocode}

\newcommand{\subsubsubsection}[1]{\paragraph{#1}\mbox{}\\}
\algrenewcommand\algorithmicrequire{\textbf{Input}:}
\algrenewcommand\algorithmicensure{\textbf{Output}:}
\algnewcommand\algorithmicdata{\textbf{Data}:}
\algblockdefx{Match}{EndMatch}% 
    [1]{\textbf{match} #1}{\textbf{end match}}%
\algblockdefx{Case}{EndCase}% 
    [1]{\textbf{case} #1}{}
\algtext*{EndCase}

\begin{document}
%-------------------------------------------------------------------------------

%don't want date printed
\date{}

\title{\Large \bf 
    BestServe: Serving Strategies with Optimal Goodput in \\
    Collocation and Disaggregation Architectures
}

%for single author (just remove % characters)
\author{
{\rm Xiannan Hu \quad\quad Tianyou Zeng \quad\quad Xiaoming Yuan}\\
The University of Hong Kong
\and
{\rm Liwei Song \quad\quad Guangyuan Zhang \quad\quad Bangzheng He}\\
Huawei Cloud, Huawei Technologies Co., Ltd.
}% end author

\maketitle

%-------------------------------------------------------------------------------
\begin{abstract}
%-------------------------------------------------------------------------------
Serving large language models (LLMs) to millions of users requires efficient resource allocation and parallelism strategies. It is a labor intensive trial-and-error process to find such a strategy.
We present \textit{BestServe}, a novel framework for ranking serving strategies by estimating goodput under various operating scenarios. Supporting both collocated and disaggregated architectures, \textit{BestServe} leverages an inference simulator built on an adapted roofline model and CPU-GPU dispatch dynamics. Our framework determines the optimal strategy in minutes on a single standard CPU, eliminating the need for costly benchmarking, while achieving predictions within a $20\%$ error margin. It appeals to be practical for rapid deployment planning because of its lightweight design and strong extensibility.
\end{abstract}

\section{Introduction} \label{sec:introduction}

Large Language Models (LLMs) such as OpenAI's o3-mini~\cite{OpenAI2025o3mini} and LLaMa 3.2~\cite{Meta2024Llama} have become increasingly accessible through cloud services~\cite{aws, azure, googlecloud, huaweicloud}. However, deploying LLMs at scale remains prohibitively expensive due to their computational demands and reliance on GPUs, a challenge described as ``eye-watering'' by OpenAI's CEO Sam Altman~\cite{Reuters2023Focus}.

State-of-the-art LLMs are predominantly decoder-only Transformer-based~\cite{10.5555/3295222.3295349} autoregressive models, with inference divided into two distinct phases. The \textit{prefill} phase generates the first token and stores ephemeral tensors as the KV-Cache for reuse in subsequent tokens, which involves intensive linear algebra operations and is commonly considered \textit{compute-bound}. The \textit{decode} phase, which autoregressively generates subsequent tokens, accesses the KV-Cache to avoid redundant computations. 
While often labeled as \textit{memory-bound}, we argue this is \textit{misleading}; the decode phase is primarily limited by CPU-GPU dispatch latency, where we relabel them as \textit{dispatch-bound}. 
Its autoregressive nature and lower computational complexity lead to underutilized computational resources, making LLM inference both time- and resource-intensive.

Serving LLMs to millions of users presents additional challenges. The throughput of LLM services depends heavily on the allocation of limited computational resources and the scheduling of prefill and decode requests, mainly due to the contrasting requirements of prefill and decode phases. Factors such as operating scenarios (e.g., patterns of requests such as sequence lengths, arrival rates, and generation lengths), hardware infrastructure, and service level objectives (SLOs) further complicate the development of effective serving strategies. To address these challenges, efficient serving engines are essential for maximizing resource utilization while maintaining low latency.

Several open-source serving engines aim to deliver high-performance LLM inference frameworks with easy access, such as NVIDIA's TensorRT-LLM \cite{nvidia-tensorrt-llm}, Hugging Face's Text Generation Inference \cite{tgi}, Orca \cite{280922}, and vLLM \cite{10.1145/3600006.3613165}. These frameworks combine existing techniques with innovative ideas to optimize inference. For instance, model and pipeline parallelism introduced in NVIDIA's Megatron-LM \cite{shoeybi2020megatronlmtrainingmultibillionparameter} are widely adopted in serving engines; Orca \cite{280922} pioneered iteration-level continuous batching; and vLLM \cite{10.1145/3600006.3613165} implemented dynamic memory management with PagedAttention. Engines like TensorRT-LLM and vLLM integrate all these techniques while supporting a wide range of models, making them highly appealing for production use. It is notable that for these engines, each GPU processes both phases of a request, leading to interference due to their contrasting natures. 

More recently, engines featuring the disaggregation of the prefill and decode phases have emerged, such as DistServe \cite{Zhong2024DistServeDP} and Mooncake \cite{qin2024mooncakekvcachecentricdisaggregatedarchitecture}. In the disaggregation architecture, GPUs are specialized for either prefill or decode tasks. Requests are first prefilled at a prefill instance, and the KV-Cache is then transmitted to a decode instance to complete the process. This specialization reduces interference and can achieve higher throughput compared to the collocated architecture in certain scenarios, though it introduces additional communication overhead when transferring requests and KV-Cache between instances.

Later in this paper, we illustrate that neither disaggregation nor collocation architecture consistently outperforms the other across all operating scenarios. Furthermore, given a specific architecture, its relative performance is highly sensitive to the ratio of prefill to decode instances and the specific operating scenarios. Current serving engines lack automated mechanisms for dynamically optimizing system architecture and configurations. Consequently, service providers must manually determine and implement optimal setups for each scenario, as this optimization critically impacts the overall performance of LLM serving systems.

However, identification the optimal configuration is generally far from straightforward. The optimal architecture and hyperparameters depend heavily on operating scenarios, available computational resources, and SLO requirements. Service providers face diverse operating scenarios due to varying user needs. For instance, summarization tasks require LLMs to process long input sequences and generate concise replies, while generation tasks involve producing lengthy outputs from short prompts. Consequently, architectures and hyperparameters must be adjusted accordingly. Additionally, SLOs must be satisfied for most requests to ensure good serving quality, imposing further constraints on the architecture and hyperparameters.

The current approach for optimizing throughput relies on trial-and-error experiments. This involves simulating real-life scenarios with dummy requests, statistically sampled arrival timestamps, and testing various configurations of architectures and hyperparameters. Each configuration requires repeated inference runs, making the process extremely time-consuming and computationally expensive, often demanding as many GPUs as needed for real deployment. While some studies \cite{Zhong2024DistServeDP} have attempted to model LLM inference using queueing theory mathematically, these efforts remain largely heuristic and lack practical applicability. The complexity of LLM inference, coupled with intricate CPU-GPU-memory interactions, makes deriving optimal configurations purely through theoretical models highly challenging, if not infeasible.

Building on the above discussions, companies seeking to efficiently serve LLMs face a critical challenge:
\begin{quotation}
    \textit{For given operating scenarios and SLO requirements, can an efficient approach be developed to select the optimal serving strategy—including both the architecture and its configuration—without relying on extensive case-by-case simulations?}
\end{quotation}

In this paper, we propose a novel analyzer called \textit{BestServe} to address the challenges of selecting optimal serving strategies for LLMs. \textit{BestServe} features a three-level hierarchical structure designed to significantly reduce computational efforts. At its core, an estimator decomposes the prefill and decode phases into atomic API calls on CPU and computational operators on GPU. Execution latencies are estimated using an adapted roofline model, while communication latencies are approximated based on the hardware's bandwidth. Notably, our model explicitly accounts for the time required to dispatch CPU instructions to GPU, a factor often being overlooked.

The rest of this paper is organized as follows. Section \ref{sec:preliminaries} provides a brief review of the necessary background knowledge on LLMs and the roofline model. Section \ref{sec:best-serve} introduces \textit{BestServe}, detailing its three-level structure and implementation. Section \ref{sec:validation} validates the performance and accuracy of \textit{BestServe} by comparing its predictions with profiled data from real tests. Section \ref{sec:limitations} discusses the limitations of \textit{BestServe} and potential areas for improvement. Finally, Section \ref{sec:conclusion} concludes the paper and outlines future research directions.

\section{Preliminaries} \label{sec:preliminaries}
This section briefly reviews some fundamental knowledge that will be useful in the sequel.

\subsection{Elements of LLM} \label{sec:elem-llm}
Modern mainstream LLMs are primarily decoder-only Transformer-based autoregressive models \cite{10.5555/3295222.3295349}. These models consist of a stack of Transformer blocks, each comprising three main components: normalization, attention, and multilayer perceptron (MLP). While the overall structure is consistent across models, different LLMs vary in the specific implementations of these components. For example, some models adopt Root Mean Square Normalization (RMSNorm) \cite{zhang2019rootmeansquarelayer} instead of LayerNorm \cite{ba2016layernormalization}, or use grouped query attention (GQA) \cite{ainslie2023gqatraininggeneralizedmultiquery} instead of multi-head attention (MHA).

In this work, we focus on the LLaMa family of models \cite{Meta2024Llama} as a representative example. However, the methodology discussed here can be generalized to other LLM architectures with appropriate modifications.

\subsection{Elements of LLM Inference}
Optimizing the inference performance of LLMs has been a persistent challenge in the field. Over the course of their development, several critical techniques have emerged that are essential to consider when simulating LLM inference. Among these, the KV-Cache and continuous batching have become indispensable components in almost every modern LLM inference pipeline. In this work, we specifically design \textit{BestServe} to ensure compatibility with and adherence to these two essential techniques.

\subsubsection{KV-Cache} \label{sec:kv-cache}
LLMs generate tokens in an autoregressive fashion, where each forward pass yields a probability distribution for the next token. To avoid redundant computations, modern LLMs employ the KV-Cache \cite{kvcache}, which stores key and value tensors during inference for reuse in subsequent tokens. This technique significantly reduces the computational complexity of the attention module by trading memory for efficiency.

The KV-Cache also differentiates the properties of the two inference phases:
\begin{itemize}
    \item \textit{prefill phase}: The LLM generates the first token and creates a cache of key and value tensors for all tokens in the initial request.
    \item \textit{decode phase}: The LLM autoregressively generates subsequent tokens using the KV-Cache to avoid redundant computations, updating the cache with the key and value tensors of the newly generated token.
\end{itemize}

\subsubsection{Continuous Batching in LLM Inference}
Continuous batching, first introduced in \cite{10.1145/3190508.3190541}, has become a standard technique in modern LLM serving engines such as Orca \cite{280922}, vLLM \cite{vllm}, and TensorRT-LLM \cite{nvidia-tensorrt-llm}. Unlike static batching, where a new batch is formed only after all requests in the previous batch are processed—leading to GPU underutilization due to the longest request in the batch—continuous batching operates at the iteration level. It dynamically inserts new requests into the batch as soon as a processing slot becomes available, significantly improving GPU utilization and reducing latency. The famous blogpost \cite{anyscale-continuous-batching} provides a detailed and vivid explanation of continuous batching.

\subsection{Serving Metrics}

The efficiency of serving is measured by \emph{throughput}, defined as the number of requests processed per second. The quality of serving is determined by adherence to service level objectives (SLOs), which typically consist of two key metrics:
\begin{itemize}
    \item \textit{time to first token (TTFT)}: Measures the latency from a request's arrival to the generation of the first token during the prefill phase.
    \item \textit{time per output token (TPOT)}: Measures the average latency between subsequent token generations during the decode phase.
\end{itemize}
A typical SLO requires that TTFT and TPOT remain below specified thresholds (e.g., 1500ms for TTFT and 70ms for TPOT), ensuring a smooth user experience .

To balance efficiency and quality, the metric \emph{goodput} has been proposed \cite{Zhong2024DistServeDP}. Goodput is defined as the maximum request arrival rate that can be served while meeting SLO attainment goals (e.g., 90\% of requests satisfying TTFT and TPOT constraints).
Goodput is favored in industry as it avoids extreme cases where high throughput compromises SLO satisfaction or strict SLO adherence results in poor throughput \cite{distserveblog}.

\subsection{Collocation and Disaggregation Architectures}

\begin{figure*}[!ht]
    \begin{subfigure}[t]{0.3875\textwidth}
        \centerline{\includegraphics[width=0.95\textwidth]{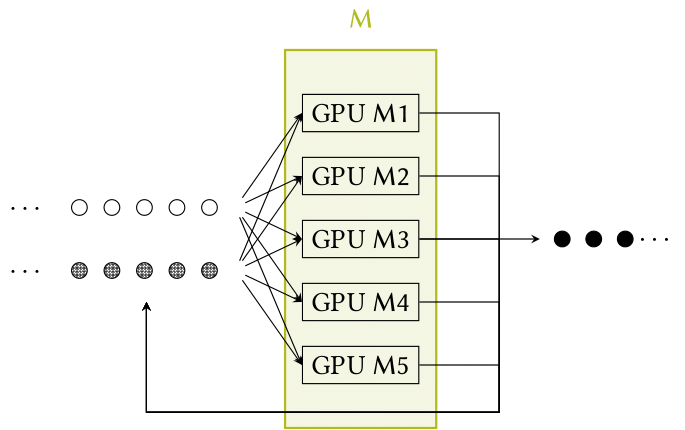}}
        \caption{Five GPU instances serving an LLM in the collocation architecture. Each GPU is a collocation instance, handling both the prefill and decode phases of requests. This setting is referred to as `5m'.}
        \label{fig:collocation}
    \end{subfigure}~\hfill~
    \begin{subfigure}[t]{0.585\textwidth}
        \centerline{\includegraphics[width=0.95\textwidth]{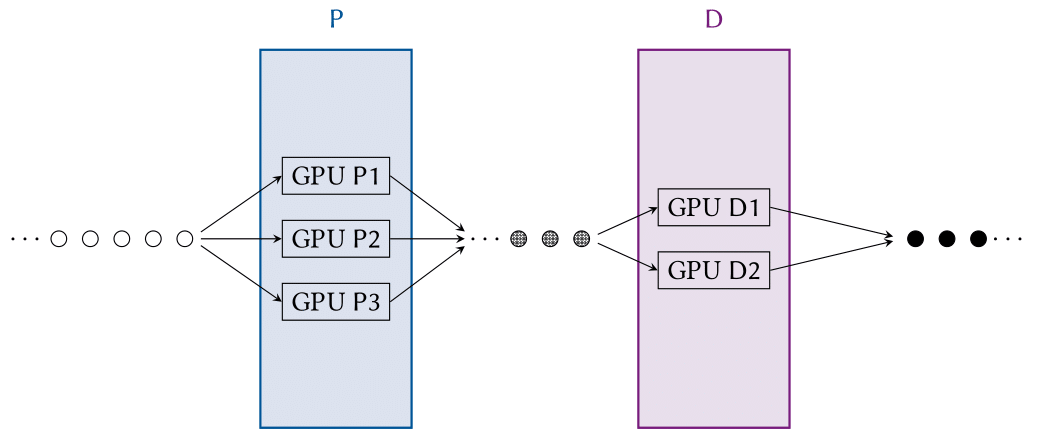}}
        \caption{Five GPU instances serving an LLM in the disaggregation architecture. Three GPU instances are specialized as prefill instances, and two GPU instances are specialized as decode instances. This setting is referred to as `3p2d'.}
        \label{fig:disaggregation}
    \end{subfigure}

    \caption{Illustration of the collocation and disaggregation architectures in LLM serving. Requests are represented by dots: hollow dots indicate new requests, gray dots indicate requests that have completed the prefill phase but not the decode phase, and black dots indicate completed requests.}
\end{figure*}

The prefill and decode phases of LLM inference, while characteristically different, share the same model weights and KV-cache. Most serving engines adopt a \emph{collocation architecture}, where GPUs or NPUs process both phases of requests on the same instance (see Figure \ref{fig:collocation}). In this architecture, batching and scheduling prefill and decode requests is challenging due to their distinct characteristics. For example, vLLM \cite{10.1145/3600006.3613165} employs a scheduler that prioritizes prefill requests and avoids batching prefill and decode requests together. While this approach optimizes TTFT, it sacrifices TPOT due to frequent interruptions of decode requests, leading to underutilization of computational resources.

An alternative approach is the \emph{disaggregation architecture}, which separates the prefill and decode phases by assigning them to specialized instances (see Figure \ref{fig:disaggregation}). Requests are first processed on prefill instances, and their KV-Cache is then transmitted to decode instances for further processing. This specialization reduces interference between the two phases but introduces additional communication overhead. Frameworks like DistServe \cite{Zhong2024DistServeDP} and Mooncake \cite{qin2024mooncakekvcachecentricdisaggregatedarchitecture} exemplify this paradigm. For consistency, we use the disaggregated serving implementation of vLLM \cite{vllm} for comparison.

To simplify notation, we use `$x$m' to denote a collocation architecture with $m$ instances and `$y$p$z$d' to denote a disaggregation architecture with $y$ prefill instances and $z$ decode instances. In this work, we focus on:
\begin{enumerate}
    \item Comparing the performance of collocation and disaggregation architectures (e.g., whether 5m outperforms 3p2d).
    \item Evaluating the impact of different prefill-to-decode ratios in the disaggregation architecture (e.g., comparing 2p3d, 3p2d, and 4p1d).
\end{enumerate}

\subsection{The Roofline Model}

The roofline model \cite{10.1145/1498765.1498785} is a widely used tool for estimating the computation time of different operators. It characterizes the performance of an operation based on its \emph{arithmetic intensity} $I$, defined as the ratio of floating-point operations (FLOPs) $W$ to memory traffic $Q$:
\begin{equation}
    I = \frac{W}{Q}.
\end{equation}

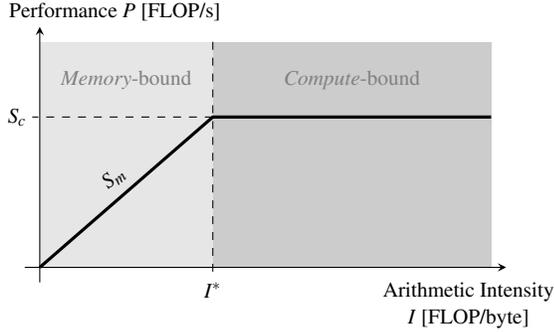
\begin{figure}[!ht]
    \centering
    \begin{tikzpicture}[scale=1, every node/.style={font=\footnotesize}]
    % \draw [help lines] (0,0) grid (6,3);
    
    \fill [gray!20] (0,0) -- (2.3,0) -- (2.3,3) -- (0,3);
    \fill [gray!40] (2.3,0) -- (2.3,3) -- (6,3) -- (6,0);

    \node [gray] at (1.15,2.5) {\textit{Memory}-bound};
    \node [gray] at (4.15,2.5) {\textit{Compute}-bound};

    \draw [-stealth] (0,-0.2) -- (0,3.2) node [shift={(1,0.2)}] {Performance $P$ [FLOP/s]};
    \draw [-stealth] (-0.2,0) -- (6.2,0) node [shift={(-0.5,-0.5)}] {\shortstack{Arithmetic Intensity \\ $I$ [FLOP/byte]} };
    \draw [very thick] (0,0) -- node [sloped, above] {$S_m$} (2.3,2) -- (6,2);
    \draw [dashed] (2.3,3) -- (2.3,-0.1) node [shift={(0,-0.2)}] {$I^*$};
    \draw [dashed] (2.3,2) -- (-0.1,2) node [shift={(-0.2,0)}] {$S_c$};
    \end{tikzpicture}
    \caption{Illustration of the roofline model.}
    \label{fig:roofline}
\end{figure}

Given the peak computation performance $S_c$ and peak memory bandwidth $S_m$ of the hardware, the roofline model states that the actual performance $\bar{P}$ of an operation is:
\begin{equation}
    \bar{P} = \min\{S_c, IS_m\}.
\end{equation}

As illustrated in Figure \ref{fig:roofline}, the model identifies a critical intensity $I^* = S_c / S_m$. Operations with $I \leq I^*$ are memory-bound, limited by $S_m$, while those with $I \geq I^*$ are compute-bound, limited by $S_c$. This framework provides a simple yet effective way to analyze the performance characteristics of various operations.

However, the original roofline model assumes ideal hardware performance, which often leads to overly optimistic estimates. To address this, we consider an \emph{adapted roofline model} that incorporates two efficiency parameters widely used in industry: the model flop utilization (MFU) $e_c$ and the model bandwidth utilization (MBU) $e_m$, where $0 < e_c, e_m \leq 1$. These parameters account for practical inefficiencies in computation and memory bandwidth.

\begin{figure}[!ht]
    \centering
    \begin{tikzpicture}[scale=1, every node/.style={font=\footnotesize}]
    % \draw [help lines] (0,0) grid (6,3);
    
    \fill [gray!20] (0,0) -- (3,0) -- (3,3) -- (0,3);
    \fill [gray!40] (3,0) -- (3,3) -- (6,3) -- (6,0);

    \node [gray] at (1.5,2.5) {\textit{Memory}-bound};
    \node [gray] at (4.5,2.5) {\textit{Compute}-bound};

    \draw [-stealth] (0,-0.2) -- (0,3.2) node [shift={(1,0.2)}] {Performance $P$ [FLOP/s]};
    \draw [-stealth] (-0.2,0) -- (6.2,0) node [shift={(-0.5,-0.5)}] {\shortstack{Arithmetic Intensity \\ $I$ [FLOP/byte]} };
    \draw [dashed] (0,0) -- node [sloped, above] {$S_m$} (2.3,2) -- (6,2);
    \draw [very thick] (0,0) -- node [sloped, below] {$e_mS_m$} (3,1.5) -- (6,1.5);
    \draw [dashed] (3,3) -- (3,-0.1) node [shift={(0,-0.2)}] {$I^*$};
    \draw [dashed] (2.3,2) -- (-0.1,2) node [shift={(-0.2,0)}] {$S_c$};
    \draw [dashed] (3,1.5) -- (-0.1,1.5) node [shift={(-0.2,0)}] {$e_cS_c$};
    \end{tikzpicture}
    \caption{Illustration of the adapted roofline model. The roofline model assumes ideal hardware performance, represented by the dashed line. The adapted model incorporates efficiency parameters: $e_m$ adjusts the slope of the memory-bound region, and $e_c$ limits the height of the compute-bound region, as shown by the solid line. This adaptation provides a more realistic estimate of performance under practical conditions.}
    \label{fig:scaled-roofline}
\end{figure}
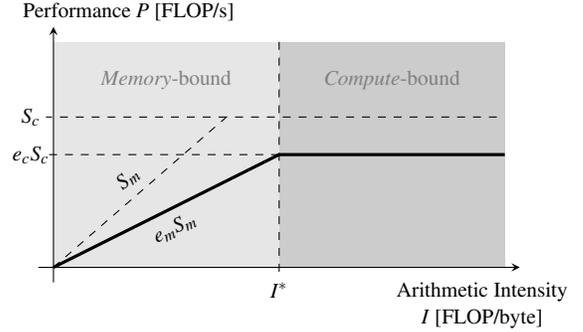

The adapted roofline model is expressed as:
\begin{equation} \label{eq:roofline}
    P = \min\{e_cS_c, Ie_mS_m\},
\end{equation}
where $P$ is the actual performance of the operation. As shown in Figure \ref{fig:scaled-roofline}, $e_m$ adjusts the slope of the roofline, while $e_c$ limits its height. This adaptation provides a more realistic estimate of performance under practical conditions.

Furthermore, the critical intensity is now influenced by the ratio of MFU to MBU:
\begin{equation} \label{eq:crit}
    I^* = \frac{e_c}{e_m} \cdot \frac{S_c}{S_m}.
\end{equation}
By substituting \eqref{eq:crit} into \eqref{eq:roofline}, we derive a simplified equivalent form:
\begin{equation}
    P = \min\{I, I^*\}e_mS_m.
\end{equation}

\begin{figure*}[!ht]
    \centering
    \includegraphics[width=\linewidth]{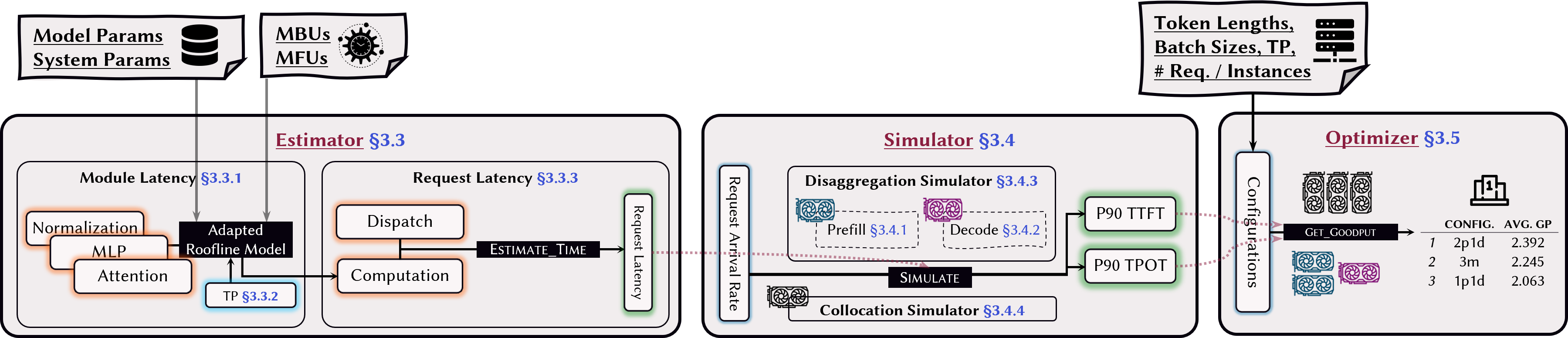}
    \caption{Structural Overview of \textit{BestServe}. The framework consists of three hierarchical components: \textit{Estimator}, which predicts operator-level latencies using an adapted roofline model; \textit{Simulator}, which models the temporal dynamics of request processing and computes serving metrics such as TTFT and TPOT; and \textit{Optimizer}, which systematically explores serving strategies to identify the configuration that maximizes goodput while satisfying SLO constraints.}
    \label{fig:structure}
\end{figure*}

This adapted roofline model forms the foundation of our \textit{Estimator} (\S \ref{sec:estimator}), enabling accurate predictions of computation time by accounting for real-world hardware constraints.

\section{BestServe} \label{sec:best-serve}

We propose \textit{BestServe}, a temporal simulator designed to address the challenge of optimizing LLM serving strategies for maximum goodput. \textit{BestServe} tackles this problem by leveraging a queueing theory-inspired interpretation of LLM inference and building upon the adapted roofline model to accurately estimate the computation time of various operators. By providing a systematic and efficient approach, \textit{BestServe} eliminates the need for labor-intensive trial-and-error experiments, enabling service providers to identify optimal configurations with minimal computational overhead.

\subsection{Motivation}

A key observation is that LLM inference in the disaggregation architecture closely resembles a tandem queue, as illustrated in Figure \ref{fig:disaggregation}. By modeling the inference process with appropriate queueing systems, we can derive many statistical insights relevant to our objectives. For instance, the throughput of the inference process can be analyzed by investigating the interdeparture rate of the queueing system, while metrics such as P90 TTFTs and TPOTs can be estimated by examining the cumulative distribution functions of queueing delays in the two stages of the tandem queue.

However, directly applying queueing theory to model LLM inference presents significant challenges. Existing analytical results are limited to relatively simple queueing systems, such as $M/M/c$ or $M/D/1$ queues, which are insufficient to capture the complexity of LLM inference. While more sophisticated queueing models may better align with the characteristics of LLM inference, their mathematical analysis often becomes intractable, leaving no clean or practical formulas to rely on.

These limitations highlight the need for a novel approach that combines the strengths of queueing theory with practical simulation techniques. \textit{BestServe} addresses this gap by leveraging a queueing theory-inspired framework while incorporating numerical simulations to overcome the analytical challenges of modeling complex LLM inference systems.

\subsection{Structural Overview}
The structural overview of \textit{BestServe} is illustrated in Figure \ref{fig:structure}. Each analysis begins with fundamental inputs, including the dimensional parameters of the model, performance specifications of the underlying hardware, and a set of fine-tuned efficiency parameters such as model flop utilization (MFU), model bandwidth utilization (MBU), and communication efficiency.

At the core of \textit{BestServe} are three hierarchical components:
\begin{itemize}
    \item \textit{Estimator} (\S \ref{sec:estimator}): This bottommost layer estimates the latency introduced by requests at the operator granularity, providing accurate predictions for both prefill and decode phases.
    \item \textit{Simulator} (\S \ref{sec:simulator}): Building on the Estimator, the Simulator models the arrival patterns of requests and performs temporal simulations to compute statistics such as arrival times, queueing delays, processing times, and departure times. The simulation logic differs slightly for the collocation architecture (\S \ref{sec:col-sim}) and the disaggregation architecture (\S \ref{sec:dis-sim}). To improve computational efficiency, the \textit{Simulator} employs a heuristic in the decode phase, later known as the pseudo batch size, which approximates continuous batching behavior while trading some simulation accuracy for significant algorithmic efficiency.
    \item \textit{Optimizer} (\S \ref{sec:optimizer}): The topmost layer enumeratively explores all possible serving strategies, leveraging the \textit{Simulator} to evaluate their goodput and identify the optimal configuration.
\end{itemize}

This hierarchical structure is designed to maximize efficiency and accuracy by dividing the workload into specialized layers. The \textit{Estimator} ensures precise computation times for individual operations, which propagates upward to enable the \textit{Simulator} to model real-world serving scenarios with high fidelity. Building on these insights, the \textit{Optimizer} systematically evaluates and compares serving strategies, eliminating the need for labor-intensive trial-and-error experiments.

By separating concerns across these layers, \textit{BestServe} achieves a balance between computational efficiency and accuracy. Its modular design allows each component to be independently improved or extended, ensuring scalability and adaptability to new hardware or LLM architectures. This approach reduces computational overhead while providing actionable insights, empowering service providers to optimize LLM serving strategies with minimal effort.

\subsection{Estimator} \label{sec:estimator}

The \textit{Estimator} forms the foundation of \textit{BestServe}, operating at the bottommost level of its hierarchical structure. Its primary role is to estimate the processing time of a batch of requests with high accuracy, as this information propagates upward to the \textit{Simulator} and \textit{Optimizer}. To achieve this, the \textit{Estimator} is equipped with an oracle (see Algorithm \ref{alg:est}) that predicts execution latencies based on the adapted roofline model.

For simplicity, we focus only on the computations within the transformer blocks introduced in \S \ref{sec:elem-llm}, as they are the primary contributors to latency. Miscellaneous operations such as embedding and tokenization are omitted, as their impact on overall latency is relatively minor and difficult to estimate accurately.

To illustrate our methodology, we focus on LLMs from the LLaMa family \cite{Meta2024Llama}, as they serve as a representative example of modern Transformer-based architectures. While our discussion centers on LLaMa models, the methodology is designed to be generalizable and can be readily adapted to other LLM architectures with appropriate modifications.

\subsubsection{Computation Time}

Each Transformer block comprises three key modules: normalization, attention, and MLP. Within each module, numerous operations are performed. For each operation, we can directly compute the amount of work and memory traffic involved. Using this information, the actual peak performance $P$ is estimated based on the adapted roofline model \eqref{eq:roofline}, providing a detailed and accurate prediction of computation time.

\begin{table}[!ht]
    \centering
    \small
    \begin{tabular}{| c | c | c | c |} 
     \hline
     $i$ & \textbf{Description}  & $W_i$ (unit: FLOP) & $Q_i$ (unit: byte) \\ 
     \hline\hline
     1 & \texttt{GATE\_PROJ} & $2bshh_0$ & $2(bs(h+h_0) + hh_0)$ \\ \hline
     2 & \texttt{SiLU} & $5bsh_0$ & $4bsh_0$ \\ \hline
     3 & \texttt{UP\_PROJ} & $2bshh_0$ & $2(bs(h+h_0) + hh_0)$ \\ \hline
     4 & \texttt{mul} & $bsh_0$ & $6bsh_0$ \\ \hline
     5 & \texttt{DOWN\_PROJ} & $2bshh_0$ & $2(bs(h+h_0) + hh_0)$ \\ \hline
     6 & \texttt{add} & $bsh$ & $4bsh_0$ \\ \hline
    \end{tabular}
    \caption{Works and memory traffics of each operation in the prefill phase of a LLaMa MLP module.}
    \label{tab:mlp-est}
\end{table}

\begin{table}[!ht]
    \centering
    \small
    \begin{tabular}{| c | c | c | c |} 
     \hline
     $i$ & \textbf{Description}  & $W_i$ (unit: FLOP) & $Q_i$ (unit: byte) \\ 
     \hline\hline
     1 & \texttt{GATE\_PROJ} & $2bhh_0$ & $2(b(h+h_0) + hh_0)$ \\ \hline
     2 & \texttt{SiLU} & $5bh_0$ & $4bh_0$ \\ \hline
     3 & \texttt{UP\_PROJ} & $2bhh_0$ & $2(b(h+h_0) + hh_0)$ \\ \hline
     4 & \texttt{mul} & $bh_0$ & $6bsh_0$ \\ \hline
     5 & \texttt{DOWN\_PROJ} & $2bhh_0$ & $2(b(h+h_0) + hh_0)$ \\ \hline
     6 & \texttt{add} & $bh$ & $4bh_0$ \\ \hline
    \end{tabular}
    \caption{Works and memory traffics of each operation in the decode phase of a LLaMa MLP module.}
    \label{tab:mlp-est-decode}
\end{table}

We start small and use the MLP module as an illustrative example to demonstrate our methodology. Each MLP module from the LLaMa family performs the following computation:
\begin{equation} 
    \mathbf{x}^i \leftarrow \mathbf{x}^i + \mathbf{P}_\downarrow (\mathrm{SiLU}(\mathbf{x}^i \mathbf{P}_\diamond) \otimes \mathbf{x}^i\mathbf{P}_\uparrow),
\end{equation}
where $P_\uparrow \in \mathbb{R}^{h \times h_0}, P_\downarrow \in \mathbb{R}^{h_0 \times h}$ and $P_\diamond \in \mathbb{R}^{h \times h_0}$ are the up, down and gate projection matrices and SiLU is the Sigmoid Linear Unit. Here, $h$ is the hidden size of attention and $h_0$ is the intermediate size of MLP.

For the prefill phase, the first operation involves a batch matrix-matrix multiplication of $\mathbf{x}^i \in \mathbb{R}^{b \times s \times h}$ and the gate projection matrix $\mathbf{P}_\diamond \in \mathbb{R}^{h \times h_0}$. Here, $b$ denotes the batch size and $s$ denotes the sequence length. This operation incurs $2bshh_0$ FLOPs. To compute it, we must read all $bsh + hh_0$ floats of parameters and write back the result, which comprises $bshh_0$ floats. Since each float is stored in \texttt{FP16} format (2 bytes per float), the lower bound of its memory traffic is estimated as $2(bs(h + h_0) + hh_0)$ bytes. Similarly, we can compute the work and memory traffic for all operations, as summarized in Table \ref{tab:mlp-est}.

Combining the work and memory traffic values from Table \ref{tab:mlp-est}, we consider the following formula to estimate the computation time of the MLP module using the adapted roofline model \eqref{eq:roofline}:
\begin{equation} \label{eq:mlp-est}
    T_\textrm{MLP} = \frac{W_1}{P_1} + \frac{W_2}{P_2} + \frac{W_3}{P_3} + \frac{W_4}{P_4} + \frac{W_5}{P_5} + \frac{W_6}{P_6},
\end{equation}
where $P_i := \min\{I_i, I^*\} e_mS_m$, and $I_i := W_i/Q_i$ for all $i = 1,2,3,4,5,6$. This formula provides a detailed breakdown of the computation time for each operation in the MLP module, ensuring accurate latency predictions.

For the decode phase, the formula \eqref{eq:mlp-est} remains applicable. However, due to the presence of the KV-Cache (as detailed in \S \ref{sec:kv-cache}), the work $W_i$ and memory traffic $Q_i$ for each operation must be adjusted to account for the reduced computational complexity and memory access patterns. The modified estimates are summarized in Table \ref{tab:mlp-est-decode}.

\begin{figure*}[!ht]
    \centering
    \begin{subfigure}[b]{\textwidth}
        \centering
        \includegraphics[width=\textwidth]{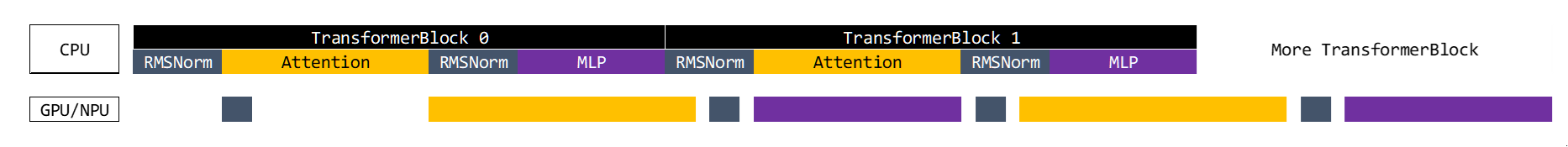}
        \caption{Compute-bound. Usually found in prefill phase.}
        \label{fig:compute-bound}
    \end{subfigure}
    \hfill
    \begin{subfigure}[b]{\textwidth}
        \centering
        \includegraphics[width=\textwidth]{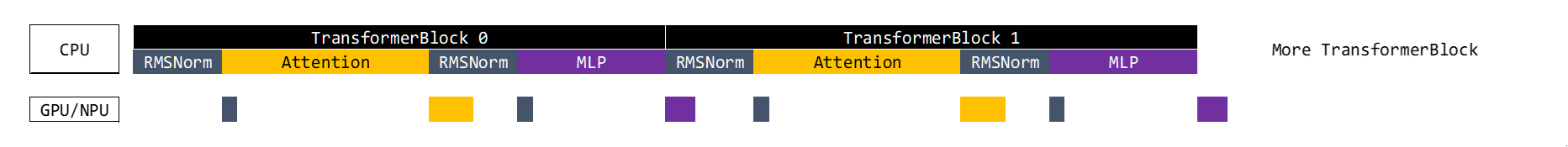}
        \caption{Dispatch-bound. Usually found in decode phase.}
        \label{fig:dispatch-bound}
    \end{subfigure}
       \caption{The two prominent patterns in LLM inference: compute-bound and dispatch-bound. For simplicity, it is assumed that each module begins computation only after all operators have been dispatched. In practice, each module contains multiple operators, and computation starts immediately after an operator is dispatched if the device is idle.}
       \label{fig:two-patterns}
\end{figure*}

A similar approach can be applied to both the normalization and attention modules to estimate their computation times. For detailed work and memory traffic calculations, as well as the corresponding formulas, we refer readers to Appendix \ref{appendix:est}.

\subsubsection{Tensor Parallelism}

To enable the deployment of extremely large LLMs, several parallelism strategies have been developed to distribute workloads across multiple devices. Among these, tensor parallelism (TP)~\cite{shoeybi2020megatronlmtrainingmultibillionparameter} is a widely adopted approach that partitions model weights and computations across devices, enabling efficient scaling of large models. While we acknowledge the existence of other parallelism schemes, such as pipeline parallelism (PP)~\cite{10.5555/3454287.3454297, harlap2018pipedreamfastefficientpipeline} and sequence parallelism (SP)~\cite{li2022sequenceparallelismlongsequence}, this work focuses on TP as a representative example. However, the generalization of our methodology to other parallelism schemes is straightforward and can be achieved with minimal modifications to the framework.

When tensor parallelism (TP) is activated, the work $W_i$ and memory traffic $Q_i$ for each operation must be adjusted based on the number of tensor replicas $t$, as the workload is distributed evenly across $t$ cards. Most terms in the work and memory traffic tables are divided by $t$, reflecting this distribution. However, certain terms, such as those involving inter-card communication or shared resources, require more careful adjustments due to their specific computation details. For a detailed breakdown of these adjustments, we refer readers to Appendix \ref{appendix:est-tp}.

On the other hand, TP incurs additional communication costs due to the need to synchronize intermediate results across cards after each attention and MLP module. These transmission costs depend on several factors, including the size of the transmitted data, the hardware configuration, the server topology, and the communication protocols used for transmitting tensors.
In practice, estimating communication costs is an ambitious and challenging task due to the numerous factors involved as elucidated above. To simplify this complexity, we adopt the following straightforward formula to approximate the transmission cost:
\begin{equation} \label{eq:trans}
    T_+ = \frac{bsh/t}{e_+S_+},
\end{equation}
where $b$ is the batch size, $s$ is the sequence length, $h$ is the hidden size, $t$ is the number of tensor replicas, $e_+$ is the communication efficiency, and $S_+$ is the hardware bandwidth. While this formula is a simplification, it captures the essential relationship between data size, parallelism, and hardware characteristics, providing a practical substitute for our purposes.

\subsubsection{Dispatch Time} \label{sec:request-latency}

Another significant contributor to the processing time of a request, often overlooked, is what we term \textit{dispatch time}. In modern computer architectures, the CPU sequentially executes instructions and issues commands to the GPU to perform computations. This process is inherently asynchronous: the GPU cannot begin computation until it receives instructions from the CPU, and it cannot proceed to the next task until the current one is completed. This asynchronous nature can introduce considerable latency, particularly in scenarios with frequent or small workloads.

\begin{algorithm*}[h!]
    \caption{Oracle for Estimating the Processing Time of a Batch of Requests}
    \label{alg:est}
    \begin{algorithmic}[1]
    \Require Batch size $b$, sequence length $s$, generation length $s_+$, tensor parallel size $t$, phase $p$, \# of Transformer blocks $\ell$.
    \Ensure Processing time $T$.
    \Procedure{estimate\_time}{$b, s, s_+, t, p, \ell$}
    \State \textcolor{red}{\texttt{@cache}} \Comment{Cache results to avoid redundant computations}
    \State \(T \gets 0\)
    \State \(T_\text{dispatch} \gets 0\), \(T_\text{compute} \gets 0\)
    \For{module in $\{\text{RMSNorm}, \text{Attention}, \text{RMSNorm}, \text{MLP}\}$}
    \State $T_\text{dispatch} \gets T_\text{dispatch} + \text{module.dispatch\_time}$ \Comment{Accumulate dispatch time}
    \If{$T_\text{dispatch} > T_\text{compute}$} \Comment{Dispatch-bound scenario}
        \State $T_\text{compute} \gets T_\text{dispatch} + \text{module.compute\_time}(b, s + s_+, t, p)$
    \Else \Comment{Compute-bound scenario}
        \State $T_\text{compute} \gets T_\text{compute} + \text{module.compute\_time}(b, s + s_+, t, p)$
    \EndIf
    \If{$t > 1$ and module.requires\_communication} \Comment{Account for communication overhead}
        \State $T_\text{compute} \gets T_\text{compute} + \text{module.communication\_time}(b, s + s_+, t, p)$
    \EndIf
    \State $T \gets T + \ell \cdot T_\text{compute}$ \Comment{Accumulate total time across layers}
    \EndFor
    \State \Return $T$
    \EndProcedure
\end{algorithmic}
\end{algorithm*}

Through extensive profiling, we identified two prominent patterns when requests are being processed:
\begin{itemize}
    \item \textit{compute-bound}: This pattern is typically observed in the prefill phase, where GPUs remain constantly busy processing queued operations dispatched from the CPU. Latency is primarily attributed to computations, as illustrated in Figure \ref{fig:compute-bound}.
    \item \textit{dispatch-bound}: This pattern is commonly found in the decode phase, where the workloads on GPUs are small. As a result, GPUs frequently remain idle, waiting for instructions from the CPU. Latency in this phase is mainly caused by the execution time of instructions and the dispatch of operations, as shown in Figure \ref{fig:dispatch-bound}.
\end{itemize}

To accurately estimate the latency incurred by a request, it is essential to account for \textit{both} the dispatch time incurred by the CPU and the computation time incurred by the GPU. Additionally, if tensor parallelism (TP) is employed, the transmission costs must also be included in the estimation. These considerations motivate the design of Algorithm \ref{alg:est}, which serves as the foundation for the \textit{Estimator}. Table \ref{tab:est-exp} illustrates how the \textit{Estimator} applies this algorithm to estimate the processing time of a single request.

\begin{table}[!ht]
    \centering
    \begin{subtable}{0.495\textwidth}
        \centering
        \begin{tabular}{lrrr}
            \toprule
            $48 \times$ & Dispatch & Compute & Communicate \\
            \midrule
            RMSNorm & 0.024 & 0.223 & 0.000 \\
            Attention & 0.190 & 2.122 & 0.100 \\
            RMSNorm & 0.024 & 0.223 & 0.000 \\
            MLP & 0.041 & 2.809 & 0.100 \\
            \midrule
            TOTAL & {\textbf{265.123}} & & \\
            \bottomrule
        \end{tabular}
        \caption{Estimated processing time for the prefill phase using Algorithm \ref{alg:est} ($b = 1$, $s = 2048$, $t = 4$, $\ell = 48$). Total time: 265.123ms.}
    \end{subtable}
    \newline
    \begin{subtable}{0.495\textwidth}
        \centering
        \begin{tabular}{lrrr}
            \toprule
            $48 \times$ & Dispatch & Compute & Communicate \\
            \midrule
            RMSNorm & 0.024 & 0.000 & 0.000 \\
            Attention & 0.190 & 0.176 & 0.100 \\
            RMSNorm & 0.024 & 0.000 & 0.000 \\
            MLP & 0.041 & 0.530 & 0.100 \\
            \midrule
            TOTAL & {\textbf{33.573}} & & \\
            \bottomrule
        \end{tabular}
        \caption{Estimated processing time for the decode phase using Algorithm \ref{alg:est} ($b = 1$, $s = 2048 + 63 = 2111$, $t = 4$, $\ell = 48$). Total time: 33.573ms.}
    \end{subtable}
    \caption{Application of Algorithm \ref{alg:est} to estimate processing times in the prefill and decode phases for CodeLlama-34b-Instruct-hf on Ascend 910B3.}
    \label{tab:est-exp}
\end{table}

In principle, the total dispatch time for each module remains consistent across all models within the same family, provided that the inference code remains unchanged. These constant dispatch times can be obtained by profiling LLM inference on smaller models (e.g., LLaMa-3.2-1B in the LLaMa family) and then applying the same constants to larger models. However, it is important to note that these figures are highly sensitive to the underlying hardware and may vary significantly across different environments.

\subsubsection{Cache by Functional Arguments}
Caching Algorithm \ref{alg:est} by its functional arguments is crucial to avoid redundant computations. This optimization significantly reduces the simulation cost, as the \textit{Simulator} frequently invokes Algorithm \ref{alg:est} with identical sets of arguments during its execution. By storing and reusing previously computed results, we ensure that the simulation process remains efficient, even when handling a large number of requests or complex scenarios.

\begin{algorithm*}[!ht]
    \caption{Simulation Routine of the Prefill Simulator}
    \label{alg:prefill-sim}
    \begin{algorithmic}[1]
    \Require A list of requests $R$, a list of arrival times $A$, maximum batch size $b_{\max}$, sequence length $s$, tensor parallel size $t$, \# of Transformer blocks $\ell$.
    \Ensure A list of departure times $D$.
    \State $T_\mathrm{current} \gets 0$, $D \gets [\inf, \inf, \ldots, \inf]$
    \State instance.when\_idle $\gets 0$ \textbf{for} instance in instances \Comment{All instances start idle}
    \While{$R \neq \varnothing$}
        \State $T_\mathrm{idle} \gets \inf$
        \For{instance in \textcolor{red}{shuffle}(instances)} \Comment{Randomize order to mimic round-robin scheduling}
            \If{instance.when\_idle $\leq T_\mathrm{current}$} \Comment{Check instance availability}
                \State $B \gets \Call{batch}{R, A, b_{\max}, T_\mathrm{current}}$ \Comment{Batch requests that have arrived by $T_\mathrm{current}$}
                \If{$B \neq \varnothing$}
                    \State $T_B \gets \Call{estimate\_time}{|B|, s, 1, t, \text{`prefill'}, \ell}$ \Comment{Estimate processing time for the batch}
                    \For{$r \in B$}
                        \State $D[r] \gets T_\mathrm{current} + T_B$ \Comment{Record departure time for each request}
                        \State Remove $r$ from $R$ \Comment{Request processed}
                    \EndFor
                    \State instance.when\_idle $\gets T_\mathrm{current} + T_B$ \Comment{Update current instance idle time}
                \EndIf
            \Else
                \State $T_\mathrm{idle} \gets \min(T_\mathrm{idle}, \text{instance.when\_idle})$ \Comment{Track the earliest time an instance becomes idle}
            \EndIf
        \EndFor
        \If{$R \neq \varnothing$}
            \State $T_\mathrm{current} \gets \max(T_\mathrm{idle}, A[R[0]])$ \Comment{Advance time to next event}
        \EndIf
    \EndWhile
    \State \Return $D$
    \end{algorithmic}
\end{algorithm*}

\subsubsection{Revisiting the Decode Phase: Dispatch-Bound, Not Memory-Bound}
We are particularly eager to correct a common misconception—that the decode phase of LLM inference is \textit{memory-bound}. In reality, this phase is primarily \textit{dispatch-bound}, with latency largely attributed to the time required for the CPU to dispatch operators to the GPU. Upgrading to hardware with higher memory bandwidth will not improve the performance of the decode phase. Instead, it is the individual operations within the decode phase that are memory-bound. The presence of the KV-Cache significantly reduces computational complexity, resulting in operations with low arithmetic intensity that are inherently memory-bound.

To mitigate dispatch-bound latency, solutions such as CUDA Graphs \cite{cudagraph} may be considered. CUDA Graphs allow the capture and replay of a sequence of GPU operations, reducing the overhead of repeatedly dispatching individual operators from the CPU to the GPU. By encapsulating multiple operations into a single dispatch, CUDA Graphs can help alleviate one of the key bottlenecks in the decode phase. While not the central focus of this work, such technologies represent a promising direction for addressing dispatch-bound latency in LLM inference.

\subsection{Simulator} \label{sec:simulator}
Building on the oracle provided by the \textit{Estimator}, the \textit{Simulator} models the temporal dynamics of LLM inference as a queueing system. Its primary purpose is to empirically compute key serving metrics, such as TTFT and TPOT, for a large number of requests under various operating scenarios. By simulating the arrival, processing, and departure of requests, the \textit{Simulator} provides insights into the performance of different serving strategies, enabling the evaluation of throughput and adherence to SLOs.

The disaggregation architecture naturally lends itself to modeling as a tandem queue, with the prefill and decode phases forming two distinct stages. This decoupling simplifies the simulation process and allows us to focus on the unique characteristics of each phase. In this section, we first describe the designs of the prefill simulator and the decode simulator, which together form the disaggregation simulator. We then explain how insights from the disaggregation architecture are adapted to model the collocation architecture, leveraging its similarities while addressing its unique challenges.

\begin{algorithm*}[h!]
    \caption{Simulation Routine of the Decode Simulator}
    \label{alg:decode-sim}
    \begin{algorithmic}[1]
    \Require A list of requests $R$, a list of arrival times $A$, maximum batch size $b_{\max}$, sequence length $s$, generation length $s_+$, tensor parallel size $t$, \# of Transformer blocks $\ell$.
    \Ensure A list of departure times $D$.
    \State $T_\mathrm{current} \gets 0$, $D \gets [\inf, \inf, \ldots, \inf]$ 
    \State instance.when\_idle[j] $\gets 0$ \textbf{for} $j$ in $\{0, 1, \ldots, b_{\max} - 1\}$ \textbf{for} instance in instances \Comment{All instances start idle}
    \While{$R \neq \varnothing$}
        \State $T_\mathrm{idle} \gets \inf$
        \For{instance in shuffle(instances)} \Comment{Randomize order to mimic round-robin scheduling}
            \If{instance.when\_idle[$j$] $\leq T_\mathrm{current}$} \Comment{Check instance availability}
                \If{$A[R[0]] < T_\mathrm{current}$}
                    \State $b \gets$ sum([instance.when\_idle[$j$] $> T_\mathrm{current}$]) \Comment{Batch size at the time of insertion}
                    \State $b^\dagger \gets \max(\lfloor (b + 1) / \tau \rfloor, 1)$ \Comment{Pseudo batch size to approximate continuous batching}
                    \State $T \gets \Call{estimate\_time}{b^\dagger, s, s_+, t, \text{`decode'}, \ell}$ \Comment{Estimate processing time for the request}
                    \State $D[R[0]] \gets T_\mathrm{current} + T$ \Comment{Record departure time for the request}
                    \State Remove $R[0]$ from $R$ \Comment{Request processed}
                    \State instance.when\_idle[$j$] $\gets T_\mathrm{current} + T$ \Comment{Update current instance idle time}
                \EndIf
            \Else
                \State $T_\mathrm{idle} \gets \min(T_\mathrm{idle}, \text{instance.when\_idle})$ \Comment{Track the earliest time an instance will have idle boxes}
            \EndIf
        \EndFor
        \If{$R \neq \varnothing$}
            \State $T_\mathrm{current} \gets \max(T_\mathrm{idle}, A[R[0]])$ \Comment{Advance time to the next event}
        \EndIf
    \EndWhile
    \State \Return $D$
    \end{algorithmic}
\end{algorithm*}

\subsubsection{Prefill Simulator} \label{sec:prefill-sim}
The simulation of prefill instances in the disaggregation architecture follows the logic outlined in Algorithm \ref{alg:prefill-sim}. The simulation progresses iteratively, advancing a time counter $T_\mathrm{current}$ until all requests are processed.

In each iteration, the simulator checks for idle prefill instances. If an idle instance is available, the simulator batches all requests that have arrived by $T_\mathrm{current}$ and assigns the batch to the instance for processing. The processing time for each batch is estimated using the \textit{Estimator}, and the departure times for all requests in the batch are recorded. The instance's idle time is then updated to reflect when it will be available again.

If no idle instances are available or no requests have arrived by $T_\mathrm{current}$, the simulator advances the time to the next event, which could be when an instance becomes idle or when a new request arrives. Once all requests are processed, Algorithm \ref{alg:prefill-sim} outputs a list of departure times, which can be used to compute percentiles of TTFTs.

We highlight the instance assignment strategy used in Algorithm \ref{alg:prefill-sim}. In practice, serving engines often adopt scheduling algorithms, such as round-robin, to dispatch requests to instances. To emulate this behavior, our approach shuffles the list of instances before each iteration. This randomization is statistically equivalent to round-robin scheduling when the number of simulated requests is large, while being simpler to implement. Additionally, Algorithm \ref{alg:prefill-sim} can be easily extended to support other scheduling schemes by implementing a dedicated function for instance selection.

\subsubsection{Decode Simulator} \label{sec:decode-sim}

The decode phase differs significantly from the prefill phase, as each request autoregressively generates multiple tokens. This introduces additional complexity in simulating the decode process. Two strategies naturally arise for decomposing and simulating the decode phase: a \emph{per token} approach or a \emph{per request} approach.

The per token approach provides a more accurate representation of continuous batching by simulating token generation iteratively. However, it requires a number of simulation loops proportional to the generation length of each request, making it computationally expensive. To balance accuracy and computational efficiency, we adopt a per request approach. This method reduces the number of simulation loops to one per request, approximating the behavior of continuous batching while significantly lowering the algorithmic complexity.

Our per request simulation strategy is detailed in Algorithm \ref{alg:decode-sim}, which provides an adequate and practical approximation for modeling the decode phase.

To quantify the degree of occupancy of each decode instance, we model them as having a fixed number of \emph{boxes}, corresponding to the prescribed maximum batch size. Instead of querying each instance for its idleness, we determine whether there are idle boxes available to batch additional requests. 

In the decode phase, each request belongs to a batch of dynamic size due to continuous batching. To approximate the time consumption of requests, we introduce the concept of a \textit{pseudo batch size} $b^\dagger$, representing the average batch size over the decode phase of each request. A naive approach is to assume that the computation of each request in the batch does not interfere with others, leading to a pseudo batch size of $b^\dagger = 1$. However, this assumption is overly optimistic and underestimates the actual workload. Conversely, taking the pseudo batch size as the actual batch size $b^\dagger = b$ at the time of insertion is overly pessimistic, as it assumes maximum interference among requests.

To balance these extremes, we propose the following heuristic formula:
\begin{align}
    b^\dagger = \max\left(
        \left\lfloor \frac{b + 1}{\tau} \right\rfloor, 1
    \right),
\end{align}
where $\tau > 0$ is a balancing scalar. This formula assumes that the effective batch size lies between the optimistic and pessimistic estimates, with $\tau$ controlling the trade-off. In practice, $\tau$ can be tuned based on the desired balance between accuracy and computational efficiency. For example, we find that $\tau = 2.5$ provides a reasonable approximation in most scenarios.

By using $b^\dagger$, we achieve a practical compromise that captures the dynamic nature of continuous batching while avoiding the computational overhead of a per token simulation. This approximation ensures that the decode simulator remains efficient and scalable, even for large-scale simulations.

\subsubsection{Disaggregation Simulator} \label{sec:dis-sim}

The output of the prefill simulator (\S \ref{sec:prefill-sim}) is a distribution of arrival times of decode requests, which is used as input for the decode simulation (\S \ref{sec:decode-sim}). By combining the prefill simulator and decode simulator, we obtain a disaggregation simulator to model the disaggregation architecture. Its primary purpose is to simulate the inference process under a given test scenario, characterized by a fixed request rate and a specific prefill-to-decode (PD) ratio. By running simulations, the disaggregation simulator computes key serving metrics, including TTFT and TPOT, providing actionable insights into the performance of different configurations.

\begin{figure}[!ht]
    \centering
    \includegraphics[width=0.5\textwidth]{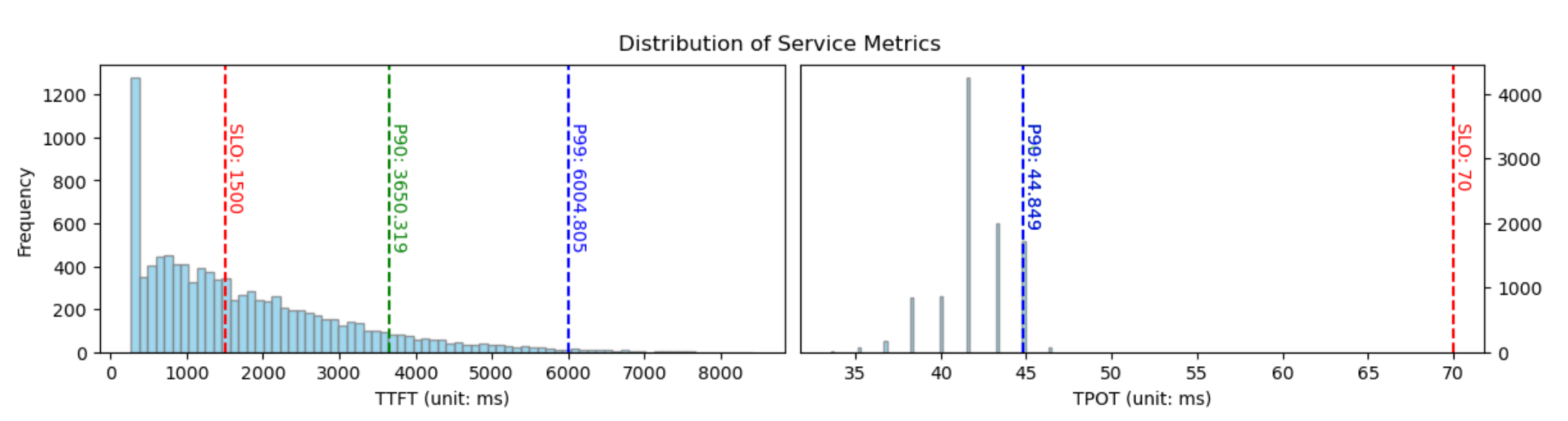}
    \caption{Visualization of service metrics from the simulation in Table \ref{tab:dis-sim-exp}. The setup corresponds to a 1p1d disaggregation scenario. The vertical axis represents the frequency, while the horizontal axes represent TTFT and TPOT. Dashed lines indicate P90 (green), P99 (blue), and SLO thresholds (red) for TTFT and TPOT.}
    \label{fig:dis-sim-vis}
\end{figure}

Table \ref{tab:dis-sim-exp} illustrates a sample output of the disaggregation simulator, showcasing its ability to estimate key metrics under a specific configuration. Additionally, the simulator can visualize the distribution of service metrics, as shown in Figure \ref{fig:dis-sim-vis}, enabling a deeper understanding of trends, bottlenecks, and system behavior.

\begin{table}[h!]
    \begin{subtable}{0.495\textwidth}
        \centering
        \begin{tabular}{lrr}
            \toprule
            Specifications & Prefill & Decode \\
            \midrule
            \# of instances & 1 & 1 \\
            Tensor parallel size & 4 & 4 \\
            Maximum batch size & 4 & 16 \\
            \midrule
            Arrival rate & 3.5 & \\
            \# of requests & 10000 & \\
            \bottomrule
        \end{tabular}
        \caption{Input specifications for the disaggregation simulator.}
        \label{tab:dis-sim-exp-setup}
    \end{subtable}
    \newline
    \begin{subtable}{0.495\textwidth}
        \centering
        \begin{tabular}{lrrr}
            \toprule
            SLO & P90 & P99 & SLO \\
            \midrule
            TTFT (ms) & 3650.319 & 6004.805 & 1500 \\
            TPOT (ms) & 44.849 & 44.849 & 70 \\
            \bottomrule
        \end{tabular}
        \caption{Estimated P90 and P99 TTFTs and TPOTs of requests based on the specifications in (a).}
    \end{subtable}
    \caption{An application of the disaggregated simulator to estimate the P90 and P99 TTFTs and TPOTs of a 1p1d scenario for CodeLlama-34b-Instruct-hf on Ascend 910B3.}
    \label{tab:dis-sim-exp}
\end{table}

\begin{figure}[!ht]
    \centering
    \includegraphics[width=0.5\textwidth]{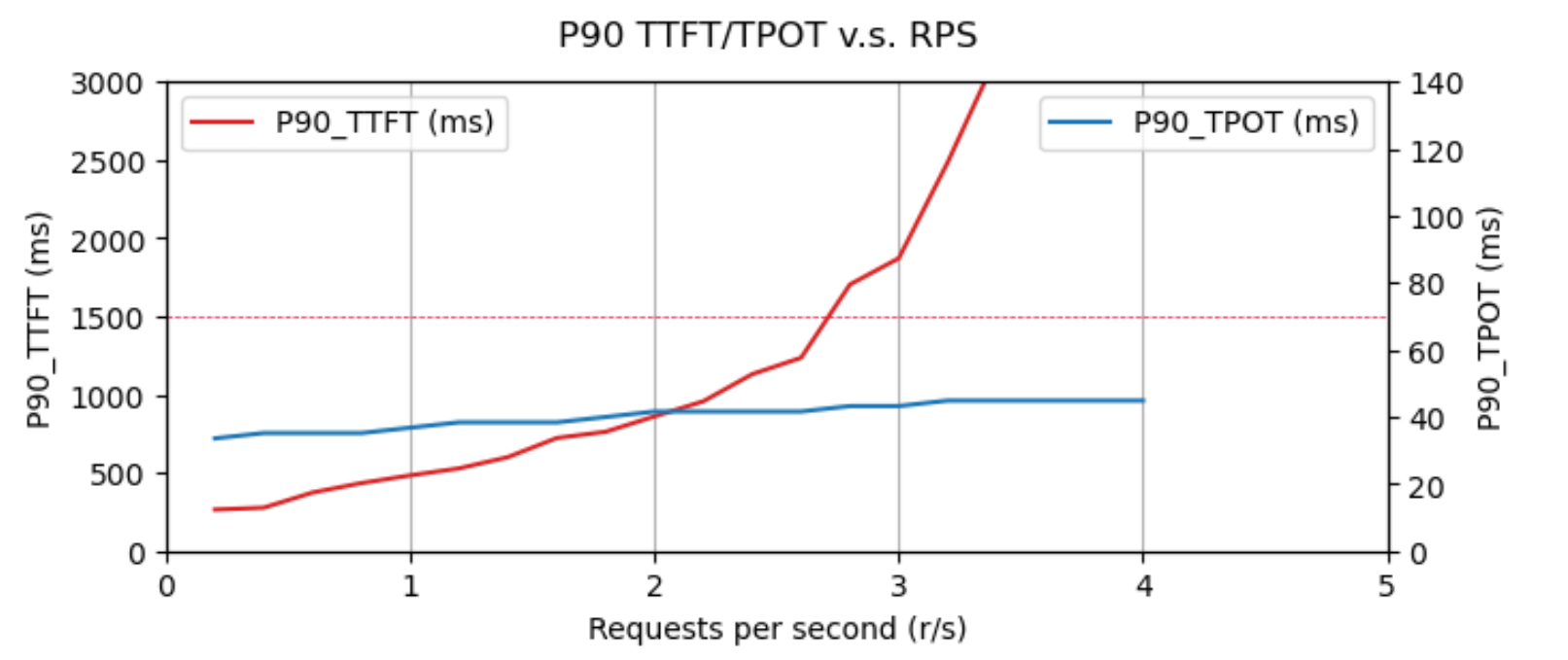}
    \caption{Performance of P90 TTFTs and TPOTs against request arrival rates. The setup corresponds to the specifications in Table \ref{tab:dis-sim-exp-setup}, with varying request arrival rates. The vertical axes represent P90 TTFT (red curve) and P90 TPOT (blue curve), while the horizontal axis corresponds to request arrival rates. The curves are interpolations of simulated P90 TTFT and TPOT at different request rates.}
    \label{fig:dis-sim-vis-slo}
\end{figure}

At this stage, goodput can be estimated by simulating the system at different request arrival rates and identifying the range where the SLOs are marginally satisfied. Figure \ref{fig:dis-sim-vis-slo} illustrates the relationship between TTFTs, TPOTs, and request arrival rates. This graph provides valuable insights for tuning efficiency parameters, enabling service providers to optimize resource allocation while maintaining adherence to SLOs.

At this stage, the goodput can be estimated by simulating the system at different request arrival rates and identifying the range where the SLOs are marginally satisfied. Figure \ref{fig:dis-sim-vis-slo} illustrates the relationship between TTFTs, TPOTs, and request arrival rates. This graph provides valuable insights for fine-tuning the hyperparameters of \textit{BestServe}, such as MFU and MBU. By analyzing the trends in the graph, users can iteratively adjust these parameters to better align the simulator's predictions with real-world performance.

\subsubsection{Collocation Simulator} \label{sec:col-sim}

When both prefill and decode of requests are performed on the same instance, the scheduling strategy will become critical to serving performance. Our collocation simulator focuses on mimicking the behavior of the current state-of-the-art vLLM \cite{10.1145/3600006.3613165} for the clarity of presentation, though the idea can be easily generalized to realize different models for various inference frameworks.

We also choose to let the collocation simulator simulate the serving of vLLM as an important baseline to set up a fair comparison between collocation and disaggregation architectures.

vLLM \cite{vllm} in particular (a) prioritizes prefills, and (b) doesn't batch prefill and decode to the same batch. Therefore, due to (b), we can treat batching of requests similarly as what we have done to the prefill simulator (\S \ref{sec:prefill-sim}) and decode simulator (\S \ref{sec:decode-sim}) for respective phase. To handle the prioritization in (a), we introduce a status flag for all instances with two states \texttt{prefill} and \texttt{decode} to indicate the nature of task the instance is currently dealing with. Just like vLLM, we will also maintain three queues corresponding to \texttt{prefill}, \texttt{decode} and \texttt{resume} requests, with the resume request signifying when suspended decode requests due to prioritization of prefills are to be resumed. The general logic of simulation is given by Algorithm \ref{alg:col-sim}.

\begin{algorithm}[h!]
    \caption{Simulation Routine of the Collocation Simulator}
    \label{alg:col-sim}
    \begin{algorithmic}[1]
    \Require A list of requests to be prefilled $P$, a list of arrival times $A$.
    \Ensure A list of prefill departure times $D_1$, a list of decode departure times $D_2$.
    \State $T_\mathrm{current} \gets 0$
    \State $D_1, D_2 \gets [\inf, \inf, \ldots, \inf]$
    \State $Q \gets \varnothing$ \Comment{Queue for decode requests}
    \State $S \gets \varnothing$ \Comment{Queue for suspended decodes}
    \While{$P \neq \varnothing$ or $Q \neq \varnothing$ or $S \neq \varnothing$}
        \State sort $S$ in ascending order of time 

        \hfill \Comment{Ensure resume requests are processed in order}
        \State next\_request\_type 
        
        \hspace{3mm} $\gets$ \Call{what\_comes\_next}{$P, Q, S, T_\mathrm{current}$}
        \Match{next\_request\_type}
            \Case{\texttt{`prefill'}}
                \State \Call{process\_prefill}{$P,Q,S,A,D_1,T_\mathrm{current}$}
            \EndCase
            \Case{\texttt{`decode'}}
                \State \Call{process\_decode}{$P,Q,D_1,D_2,T_\mathrm{current}$}
            \EndCase
            \Case{\texttt{`resume'}}
                \State idx $\gets S$.popleft().instance\_idx
                \State instances[idx].status $\gets$ decode
            \EndCase
        \EndMatch
    \EndWhile
    \State \Return $D_1, D_2$
    \end{algorithmic}
\end{algorithm}

\subsubsubsection{Availability Logic}
The processing of prefill and decode events in the collocation simulator follows similar procedures to those in Algorithms \ref{alg:prefill-sim} and \ref{alg:decode-sim}. However, additional logic is required to handle the intricacies of prioritization and suspension. Specifically, when determining the availability of an instance for either prefill or decode requests, both the current status of the instance and the nature of the next event must be considered, as outlined in Algorithm \ref{alg:available}.
\begin{algorithm}[h!]
    \caption{Availability of a Collocation Instance}
    \label{alg:available}
    \begin{algorithmic}[1]
    \Require Nature of the next request \texttt{next\_request\_type}.
    \Ensure A boolean variable indicating availability of this instance for the incoming request.
    \Procedure{idle}{instance.status, \texttt{next\_request\_type}}
    \Match{[instance.status, \texttt{next\_request\_type}]}
        \Case{\texttt{[`prefill', `prefill']}}
            \State \Return instance.when\_idle\_prefill $\leq T_\mathrm{current}$
        \EndCase
        \Case{\texttt{[`decode', `decode']}}
            \State \Return any([
            \State \hspace{1.5em} instance.when\_idle[$j$] $\leq T_\mathrm{current}$
            \State \hspace{1.5em} \textbf{for} all $j$
            \State ])
        \EndCase
        \Case{\texttt{[`decode', `prefill']}} 
        \State \hfill \Comment{Prefill prioritization}
            \State \Return True
        \EndCase
        \Case{\texttt{[`prefill', `decode']}}
            \State \Return instance.when\_idle\_prefill $\leq T_\mathrm{current}$ 
            \State \textbf{and} any([
            \State \hspace{1.5em} instance.when\_idle[$j$] $\leq T_\mathrm{current}$ 
            \State \hspace{1.5em} \textbf{for} all $j$
            \State ])
        \EndCase
    \EndMatch
    \EndProcedure
    \end{algorithmic}
\end{algorithm}

When the status of the instance and the incoming event are of the same type, determining the availability of the instance follows the same logic as in Algorithms \ref{alg:prefill-sim} and \ref{alg:decode-sim}. However, additional considerations are required when the instance's current task and the incoming event differ.

If the instance is performing decode requests and a prefill event arrives, Algorithm \ref{alg:available} always returns \texttt{True}, as prefill requests are prioritized. In this case, the instance suspends all ongoing decode requests to process the prefill request. The suspended decode requests are added to the \texttt{resume} queue, ensuring they are resumed once the prefill task is completed.

\algrenewcommand\algorithmicdata{\textbf{Data}:}
\algnewcommand\Data{\item[\algorithmicdata]}%
\begin{algorithm*}[h!]
    \caption{Processing of Prefill Requests in the Collocation Simulator}
    \label{alg:col-sim-prefill}
    \begin{algorithmic}[1]
    \Data Maximum batch size $b_{\max}$, sequence length $s$, tensor parallel size $t$, \# of Transformer blocks $\ell$.
    \Require A list of prefill requests $P$, a list of decode requests $Q$, a list of resume requests $S$, a list of arrival times $A$, a list of prefill departure times $D_1$, current time $T_\mathrm{current}$.
    \Procedure{process\_prefill}{$P,Q,S,A,D_1,T_\mathrm{current}$}
    \State $T_\mathrm{idle} \gets \inf$ 
    \For{instance in shuffle(instances)} \Comment{Randomize order to mimic round-robin scheduling}
        \If{\textcolor{blue}{instance.\Call{idle}{instance.status, \texttt{`prefill'}}}} \Comment{Check instance availability}
            \State $B \gets \Call{batch}{P, A, b_{\max}, T_\mathrm{current}}$ \Comment{Batch prefill requests}
            \If{$B \neq \varnothing$}
                \State $T_B \gets \Call{estimate\_time}{|B|, s, 1, t, \texttt{`prefill'}, \ell}$ \Comment{Estimate processing time for the batch}
                \For{$r \in B$}
                    \State $D_1[r] \gets T_\mathrm{current} + T_B$ \Comment{Record departure time for each request}
                    \State Move $r$ from $P$ to $Q$ \Comment{Move prefilled requests to decode queue}
                \EndFor
                \State instance.when\_idle\_prefill $\gets T_\mathrm{current} + T_B$ \Comment{Update current instance idle time}
                \color{blue}
                \Match{instance.status} \Comment{Handle suspension or delay of decode requests}
                \Case{\texttt{`decode'}}
                    \State instance.status $\gets$ \texttt{`prefill'} \Comment{Prefill prioritization}
                    \State $S$.append(instance.suspend\_decode) \Comment{Suspend ongoing decodes}
                \EndCase
                \Case{\texttt{`prefill'}}
                    \State $S$.update(instance.delay\_decode) \Comment{Further postpone decode}
                \EndCase
                \EndMatch
                \color{black}
                \State \textbf{break} \Comment{Process once and exit}
            \EndIf
        \Else
            \State $T_\mathrm{idle} \gets \min(T_\mathrm{idle}, \text{instance.when\_idle})$ \Comment{Track the earliest idle time of any instance}
        \EndIf
    \EndFor
    \color{blue}
    \State $T_\mathrm{next} \gets$ $T_\text{\Call{what\_comes\_next}{$P,Q,S$}}$ \Comment{Determine time of next event}
    \color{black}
    \State $T_\mathrm{current} \gets \max(T_\mathrm{idle}, T_\mathrm{next})$ \Comment{Advance time to the next event}
    \EndProcedure
    \end{algorithmic}
\end{algorithm*}

Conversely, if the instance is performing prefill requests and a decode event arrives, availability is determined by two conditions: (1) no prefill requests are currently being processed, and (2) there is sufficient capacity to batch additional decode requests. These checks ensure that decode requests are only processed when the instance is idle and no prioritized prefill has yet been finished.

\subsubsubsection{Advancement Logic} 
In addition to the unique logic for determining instance availability, advancing time in the simulation requires a more holistic approach. Specifically, when an attempt to process either prefill or decode requests fails, the simulation must advance the time to the occurrence of the next event. This event could correspond to the arrival of a new request or the availability of an instance, regardless of the request type.

\subsubsubsection{Processing Prefill and Decode}
The procedures for processing prefill and decode events in the collocation simulator are detailed in Algorithms \ref{alg:col-sim-prefill} and \ref{alg:col-sim-decode}, respectively. These algorithms are adapted from Algorithms \ref{alg:prefill-sim} and \ref{alg:decode-sim} to account for the unique characteristics of the collocation architecture. Key enhancements, such as handling prioritization and suspension of decode requests, are highlighted in blue for better clarity.

\begin{algorithm*}[h!]
    \caption{Processing of Decode Requests in the Collocation Simulator}
    \label{alg:col-sim-decode}
    \begin{algorithmic}[1]
    \Data Maximum batch size $b_{\max}$, sequence length $s$, generation length $s_+$, tensor parallel size $t$, \# of Transformer blocks $\ell$.
    \Require A list of prefill requests $P$, a list of decode requests $Q$,  a list of prefill departure times $D_1$, a list of decode departure times $D_2$, current time $T_\mathrm{current}$.
    \Procedure{process\_decode}{$P,Q,D_1,D_2,T_\mathrm{current}$}
    \State $T_\mathrm{idle} \gets \inf$ 
    \For{instance in shuffle(instances)} \Comment{Randomize order to mimic round-robin scheduling}
        \If{\textcolor{blue}{instance.\Call{idle}{instance.status, `decode'}}} \Comment{Check instance availability}
            \If{$D_1[Q[0]] < T_\mathrm{current}$}
                \State $b \gets$ sum([instance.when\_idle[$j$] $> T_\mathrm{current}$]) \Comment{Batch size at the time of insertion}
                \State $b^\dagger \gets \max(\lfloor (b + 1) / \tau \rfloor, 1)$ \Comment{Pseudo batch size to approximate continuous batching}
                \State $T \gets \Call{estimate\_time}{b^\dagger, s, s_+, t, \text{`decode'}, \ell}$ \Comment{Estimate processing time for the request}
                \State $D_2[Q[0]] \gets T_\mathrm{current} + T$ \Comment{Record departure time for the request}
                \State Remove $Q[0]$ from $Q$ \Comment{Request processed}
                \State instance.when\_idle[$j$] $\gets T_\mathrm{current} + T$
                \State \textbf{break} \Comment{Process once and exit}
            \EndIf
        \Else
            \State $T_\mathrm{idle} \gets \min(T_\mathrm{idle}, \text{instance.when\_idle})$ \Comment{Update current instance idle time}
        \EndIf
    \EndFor
    \color{blue}
    \State $T_\mathrm{next} \gets$ \Call{when\_next\_request}{$P,Q,S$} \Comment{When is the next request coming}
    \color{black}
    \State $T_\mathrm{current} \gets \max(T_\mathrm{idle}, T_\mathrm{next})$ \Comment{Advance if no idle instance or no request arrived}
    \EndProcedure
    \end{algorithmic}
\end{algorithm*}

\subsubsubsection{Handling Prefill Prioritization}
When processing prefill requests, we consider two important scenarios to properly handle the prioritization of prefills and the suspension of decodes:
\begin{itemize}
    \item If the instance is in the \texttt{decode} state, ongoing decode requests are suspended to prioritize the incoming prefill request. The time at which the decode requests can be resumed is recorded and constructed as a resume request, as reflected in Lines 14-16 of Algorithm \ref{alg:col-sim-prefill}.
    \item If the instance is in the \texttt{`prefill'} state, consecutive prioritization of prefill requests causes further delays in the resumption of suspended decode requests on this instance. As a result, the corresponding resume request must be updated, as reflected in Lines 17-18 of Algorithm \ref{alg:col-sim-prefill}. This is also why the resume queue $S$ needs to be constantly resorted in Algorithm \ref{alg:col-sim}, ensuring that suspended decode requests are resumed in the correct order.
\end{itemize}

\begin{table}[h!]
    \begin{subtable}{0.495\textwidth}
        \centering
        \begin{tabular}{lr}
            \toprule
            Specifications & Collocation \\
            \midrule
            \# of instances & 2 \\
            Tensor parallel size & 4 \\
            Maximum batch size & 4 \\
            \midrule
            Arrival rate & 3.5 \\
            \# of requests & 10000 \\
            \bottomrule
        \end{tabular}
        \caption{Input specifications for the collocation simulator.}
        \label{tab:col-sim-exp-setup} 
    \end{subtable}
    \newline
    \begin{subtable}{0.495\textwidth}
        \centering
        \begin{tabular}{lrrr}
            \toprule
            SLO & P90 & P99 & SLO \\
            \midrule
            TTFT (ms) & 556.309 & 1091.503 & 1500 \\
            TPOT (ms) & 4360.659 & 4656.043 & 70 \\
            \bottomrule
        \end{tabular}
        \caption{Estimated P90 and P99 TTFTs and TPOTs of requests based on the specifications in (a).}
    \end{subtable}
    \caption{An application of the collocation simulator to estimate the P90 and P99 TTFTs and TPOTs of a 2m scenario for CodeLlama-34b-Instruct-hf on Ascend 910B3.}
    \label{tab:col-sim-exp}
\end{table}

\subsubsubsection{Decode Resumption}
The resumption of unfinished decode requests is straightforward. It involves flipping the status of the instance from \texttt{prefill} to \texttt{decode}, as shown in Lines 14-15 of Algorithm \ref{alg:col-sim}. This simple operation ensures that suspended decode requests are resumed seamlessly once the instance completes its prefill tasks.

\begin{figure}[th!]
    \centering
    \includegraphics[width=0.5\textwidth]{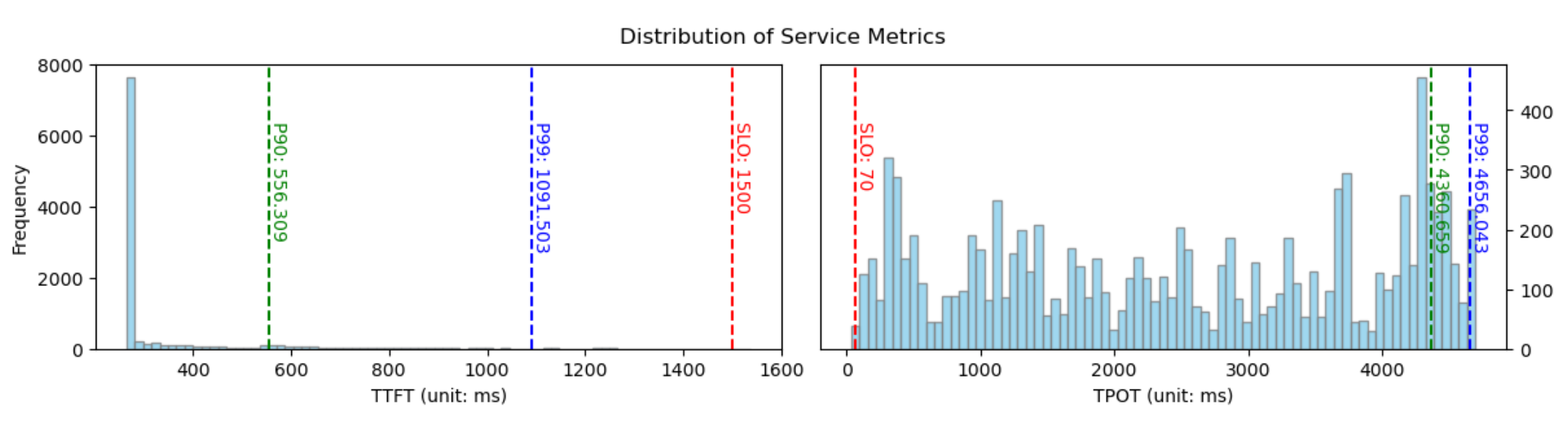}
    \caption{Distribution of service metrics from the simulation in Table \ref{tab:col-sim-exp}. The setup corresponds to a 2m collocation scenario. The vertical axis represents the frequency, while the horizontal axes represent TTFT and TPOT. Dashed lines indicate P90 (green), P99 (blue), and SLO thresholds (red) for TTFT and TPOT.}
    \label{fig:col-sim-vis}
\end{figure}

\begin{figure}[th!]
    \centering
    \includegraphics[width=0.5\textwidth]{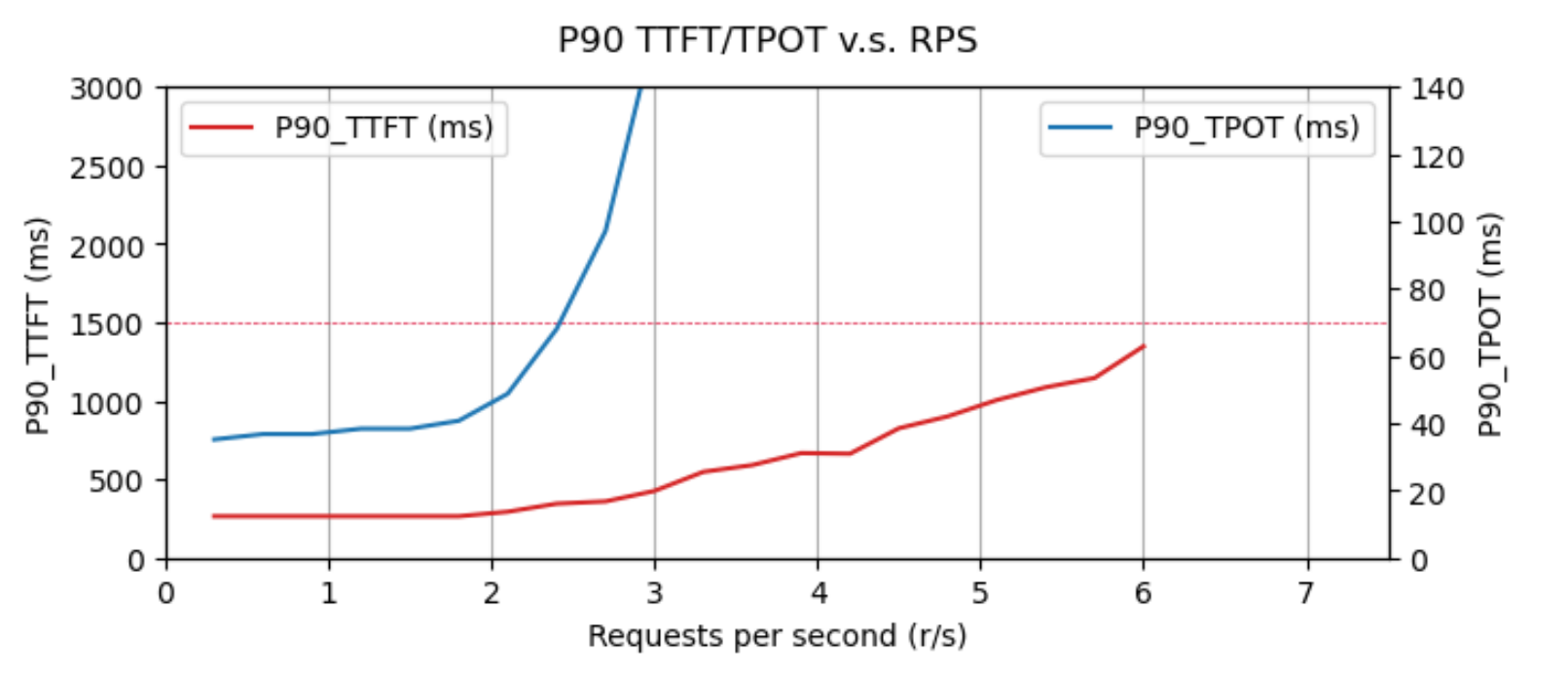}
    \caption{Performance of P90 TTFTs and TPOTs against request arrival rates. The setup corresponds to the specifications in Table \ref{tab:col-sim-exp-setup}, with varying request arrival rates. The vertical axes represent P90 TTFT (red curve) and P90 TPOT (blue curve), while the horizontal axis corresponds to request arrival rates. The curves are interpolations of simulated P90 TTFT and TPOT at different request rates.}
    \label{fig:col-sim-vis-slo}
\end{figure}

Table \ref{tab:col-sim-exp} demonstrates the capability of the collocation simulator to estimate key metrics under a specific configuration. Similar to the disaggregation simulator, the collocation simulator also provides visualizations of service metrics, as shown in Figure \ref{fig:col-sim-vis}. We continue to have Figure \ref{fig:col-sim-vis-slo} illustrating the relationship between TTFTs, TPOTs, and request arrival rates, useful when fine-tuning efficiency parameters.


\subsection{Optimizer} \label{sec:optimizer}

The overarching component of \textit{BestServe} is the \textit{Optimizer}, which focuses on enumerating all permissible configurations and determining their corresponding goodputs. It leverages the \textit{Simulator} as the primary tool to evaluate the performance of each configuration under various operating scenarios.
 
Users are required to specify the following set of parameters to define the test ground for evaluating and comparing serving strategies:
\begin{enumerate}[noitemsep]
    \item The sequence length $s$ and generation length $s_+$ of requests to be simulated,
    \item The total number of requests to simulate,
    \item The maximum number of instances to consider in both architectures,
    \item A set of admissible tensor parallel sizes for instances in both architectures, and
    \item A \textit{fixed} maximum batch size for instances in both architectures.
\end{enumerate}
These parameters provide a standardized framework for testing, ensuring that all serving strategies are evaluated under consistent and comparable conditions.

\begin{algorithm}[!ht]
\caption{Finding the Goodput of a Serving Strategy}
\label{alg:bisect}
\begin{algorithmic}[1]
    \Require A serving strategy $\mathcal{S}$, a tolerance level $\epsilon$.
    \Ensure Goodput of this serving strategy $\lambda_\mathcal{S}$.
    \Procedure{get\_goodput}{$\mathcal{S}$}
    \State $\lambda_\ell = 0.1$
    \State $\lambda_u = 1.2 / T_{min}$
    
    \Comment{$T_{min}$: min. time to process one request under $\mathcal{S}$}
    \If{\textbf{not} \Call{feasible}{$\lambda_\ell$}}
    \State \Return 0
    \EndIf
    \While{$\lambda_u - \lambda_\ell < \epsilon$}
    \State $\lambda_m = (\lambda_\ell + \lambda_u) / 2$
    \If{\Call{feasible}{$\lambda_m$}}
    \State $\lambda_\ell = \lambda_m$
    \Else
    \State $\lambda_u = \lambda_m$
    \EndIf
    \EndWhile
    \State \Return $\lambda_\mathcal{S} = \lambda_{\ell}$
    \EndProcedure
    \end{algorithmic}
\end{algorithm}

At its core, the \textit{Optimizer} employs a bisection-based method to determine the goodput for each serving strategy, as detailed in Algorithm \ref{alg:bisect}. This method uses two pointers, $\lambda_\ell$ and $\lambda_u$, to represent the lower and upper bounds of goodput, measured in requests per second.

The initial lower bound, $\lambda_\ell = 0.1$, is chosen as a pessimistic arrival rate. Any serving strategy unable to meet this rate is immediately rejected. The initial upper bound, $\lambda_u = 1.2/T_{\min}$, is inspired by queueing theory \cite{10.5555/1972549}. Here, $T_{\min}$ represents the minimum time required to process a single request under the given serving strategy, as estimated by the \textit{Estimator}. The value of $\lambda_u$ is set slightly higher than the optimistic serving rate $1/T_{\min}$, as exceeding this rate would cause the average queueing delay to grow indefinitely, violating SLO constraints in the long run.

The main body of the algorithm is as mentioned based on a bisection method. The goal is to maintain $\lambda_\ell$ as a request rate that satisfies the SLO constraints, while narrowing the gap between the lower bound $\lambda_\ell$ and the upper bound $\lambda_u$. Feasibility checks are performed using the \textit{Simulator}, as detailed in Algorithm \ref{alg:feasible}. The bisection process continues until the difference between the two bounds is sufficiently small, ensuring an accurate estimate of the maximum goodput.

\begin{algorithm}[!ht]
\caption{Adherence to P90 SLO}
\label{alg:feasible}
\begin{algorithmic}[1]
    \Require A request rate $\lambda$.
    \Ensure A boolean variable indicating whether the SLO constraints can be satisfied under this request rate.
    \Procedure{feasible}{$\lambda$}
    \State $\tau := 0.1$ \Comment{relaxation factor}
    \State \texttt{P90\_TTFT, P90\_TPOT} $\gets$ \Call{simulate}{$\lambda$}
    \State \Return \phantom{\textbf{and}} \texttt{P90\_TTFT} $\leq (1+\tau)$\texttt{TTFT\_GOAL}

    \quad\quad\; \textbf{and} \texttt{P90\_TPOT} $\leq (1+\tau)$\texttt{TPOT\_GOAL}
    \EndProcedure
\end{algorithmic}
\end{algorithm}

A simulator is invoked to model LLM inference on the specified test ground. The resulting simulated P90 TTFT and P90 TPOT are then compared against the SLO goals, with a relaxation factor $\tau = 0.1$. This factor is not chosen arbitrarily; it plays a crucial role in addressing the intrinsic stochasticity of request arrivals, which are modeled as a Poisson process in this work. Even under identical test conditions, the simulated P90 TTFT and P90 TPOT can exhibit significant variability due to this stochastic nature. Setting a hard feasibility check with $\tau = 0$ in Algorithm \ref{alg:feasible} would result in an underestimated goodput for all serving strategies, as it fails to account for this inherent variability. 

\begin{figure}[!ht]
    \begin{subfigure}[c]{0.45\textwidth}
        \centerline{\includegraphics[width=0.99\textwidth]{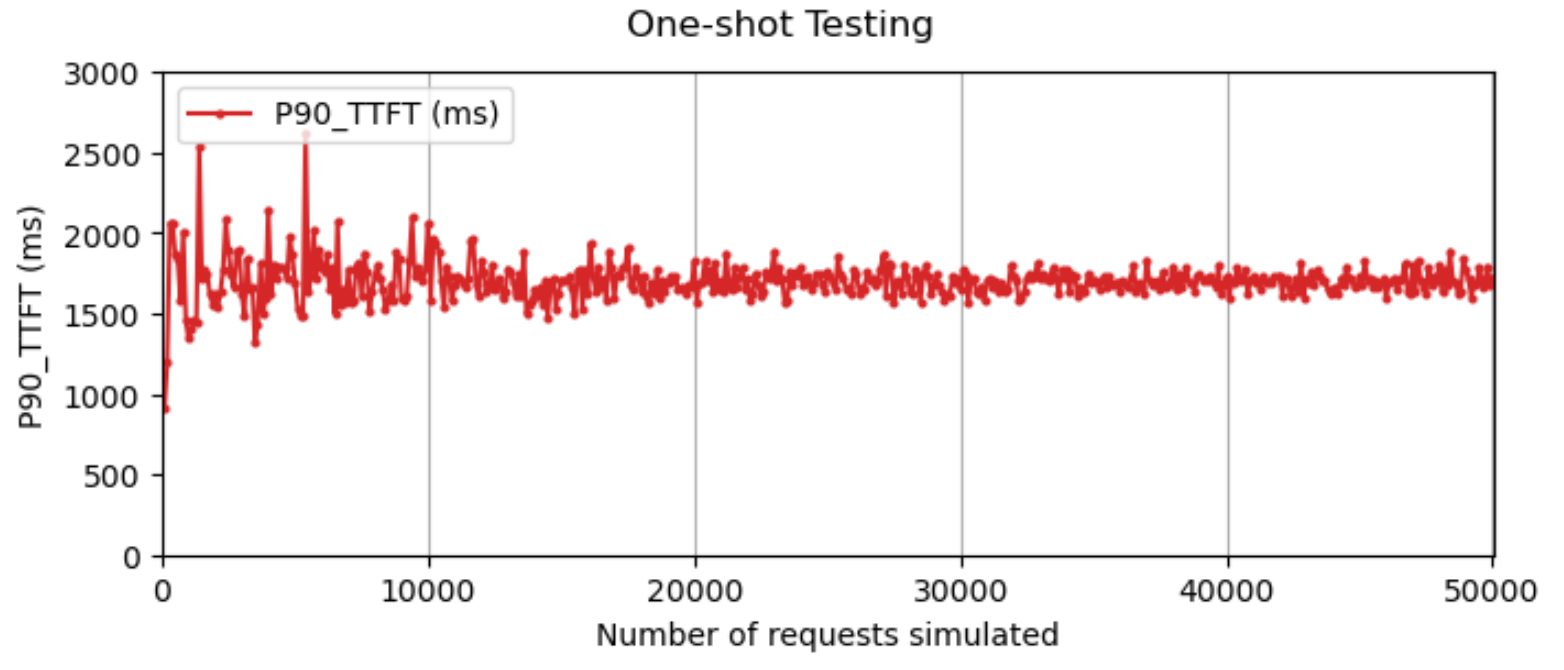}}
        \caption{One-shot simulation. Variance in P90 TTFTs remains non-negligible, even when the number of requests simulated is exceedingly large. This highlights the limitations of relying solely on one-shot testing.}
        \label{fig:one-shot}
    \end{subfigure}

    \begin{subfigure}[c]{0.45\textwidth}
        \centerline{\includegraphics[width=0.99\textwidth]{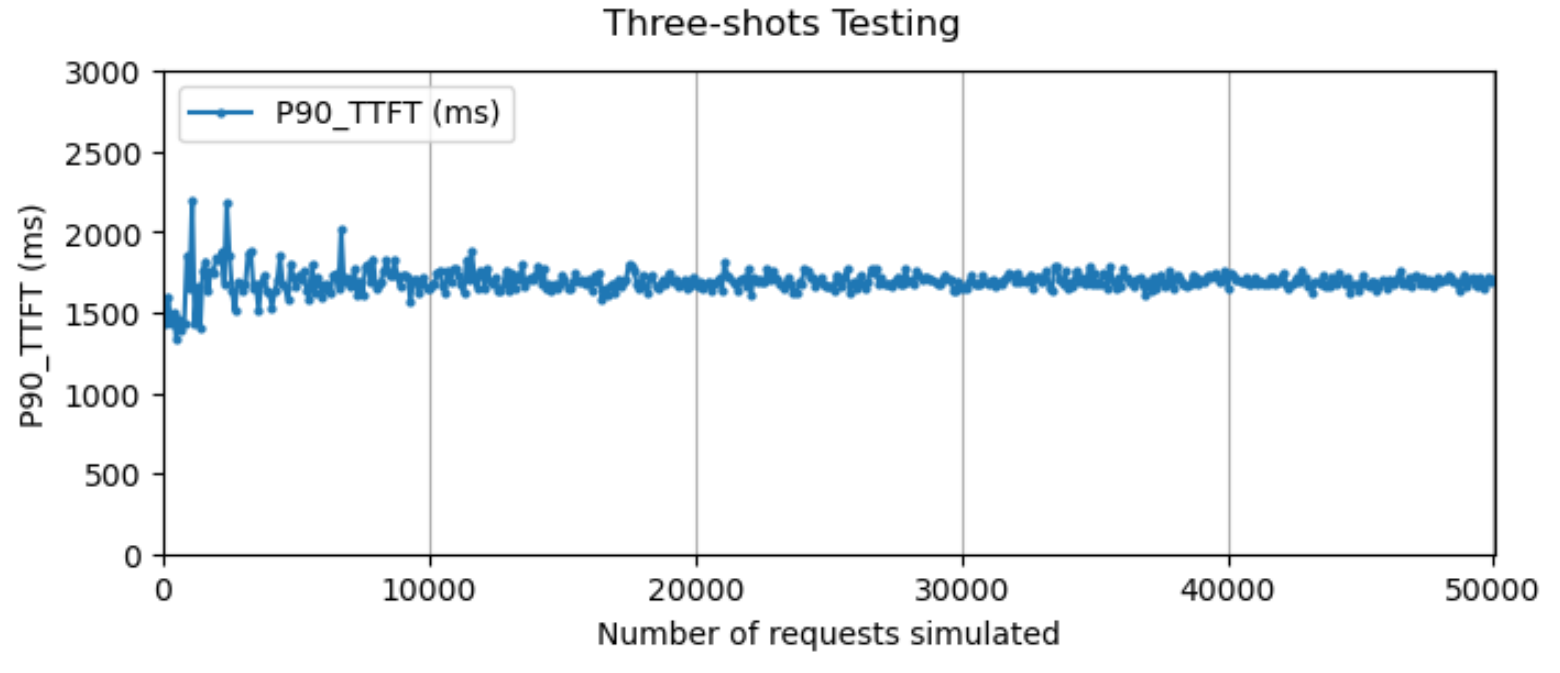}}
        \caption{Repetitive testing and averaging. Simulations are carried out three times, and the P90 TTFT is reported as the average across the three runs. Reduction of variance and more consistent results compared to one-shot testing are observed.}
        \label{fig:few-shot}
    \end{subfigure}

    \caption{Impact of one-shot and repetitive testing on the variance of simulated P90 TTFTs. The vertical axis represents the P90 TTFT, while the horizontal axis corresponds to the number of requests simulated.}
    \label{fig:shots}
\end{figure}

Figure \ref{fig:one-shot} illustrates that even when the number of requests simulated is exceedingly large—far beyond what would realistically be simulated—the P90 TTFT still oscillates within a range of approximately $\pm 5\%$. This variability underscores the necessity of introducing a relaxation factor to ensure robust goodput estimation despite the stochastic fluctuations in the simulation.

While we choose $\tau = 0.1$ in this work, motivated by our empirical observation in Figure \ref{fig:one-shot}, we acknowledge that this choice is not well-educated. We do wish to invite experts in related fields to provide guidance on making more educated choices for $\tau$ based on statistical principles.

Another observation worth noting is that repetitive testing and averaging the results can reduce variance to some extents, as demonstrated in Figure \ref{fig:few-shot}. Although not the primary focus of this work, this finding highlights the limitations of one-shot testing, especially given the persistent variance observed even when the number of requests simulated is large. This serves as a reminder for practitioners performing real inference tests: relying solely on one-shot testing may lead to misleading conclusions due to the inherent variability in the results. Repeated testing and averaging might provide a more reliable basis for evaluating serving strategies.

\section{Validation} \label{sec:validation}

To evaluate the effectiveness and accuracy of \textit{BestServe}, we conduct a series of experiments comparing its predictions with ground truth obtained from manual benchmarking. The validation focuses on assessing how well \textit{BestServe} can estimate the maximum goodput of various serving strategies under different operating scenarios. By leveraging a high-performance computing (HPC) testbed and a representative large language model (LLM), we demonstrate that \textit{BestServe} achieves reasonable accuracy with significantly reduced computational overhead. This section outlines the experimental setup, implementation details, and a comprehensive comparison of \textit{BestServe}'s predictions against manually profiled results, highlighting its practicality and efficiency for early-stage deployment planning.

\subsection{Setup}

\textbf{Testbed.}\; Our testbed consists of a high-performance computing (HPC) node cluster. Each node in the cluster is equipped with eight Ascend 910B3 NPUs, each capable of delivering 313 TFLOPs of computational power. The nodes are interconnected via three HCCS links, providing a maximum bandwidth of 90 GB/s. While our experiments are conducted on NPUs, the results remain applicable to GPU-based systems. This is because the CPU-NPU dynamics, such as dispatch latency and memory bandwidth utilization, are fundamentally similar to CPU-GPU interactions. Therefore, the insights and conclusions drawn from this study are equally valid for GPU-based serving strategies.

\noindent
\textbf{Model.}\; For our evaluation, we utilize the CodeLlama-34b-Instruct-hf model, a mid-sized large language model (LLM) from the LLaMa family. This model is selected because it is sufficiently large to stress the computational resources of the testbed while remaining manageable for real inference tests with longer sequence lengths.

\noindent
\textbf{Metric.}\; We use normalized goodput as the primary evaluation metric, defined as the goodput divided by the number of GPUs used in the serving strategy. This metric provides a clear and fair comparison of resource utilization across different serving strategies.

\noindent
\textbf{Ground truth.}\; To establish ground truth, we use vLLM-Ascend \cite{vllm-ascend} as the serving platform. Dummy requests are generated with fixed input sequence lengths, generation lengths, and arrival times sampled from a Poisson process to mimic real-world request patterns. We profile LLM inference on these dummy requests across various request rates and serving strategies under four operating scenarios. Due to the labor-intensive nature of manual benchmarking, we approximate the maximum goodput by testing a limited number of request rates for each strategy. Line charts of P90 TTFTs and P90 TPOTs against request rates, similar to Figures \ref{fig:dis-sim-vis-slo} and \ref{fig:col-sim-vis-slo}, are used to approximately identify the highest request rate that satisfies both SLOs. This value is then used as the ground truth.

\noindent
\textbf{Operating Scenarios.}\; Although \textit{BestServe} is designed to handle variable-length requests, this evaluation focuses on fixed input and output sequence lengths to eliminate variability introduced by stochastic request lengths. Four operating scenarios are considered:
\begin{itemize}[noitemsep]
    \item OP1: Input sequence length 8192, generation length 512,
    \item OP2: Input sequence length 2048, generation length 64,
    \item OP3: Input sequence length 1024, generation length 64,
    \item OP4: Input sequence length 256, generation length 2048.
\end{itemize}

\noindent
\textbf{Hyperparameters of BestServe.}\; The accuracy of \textit{BestServe}'s predictions relies on several efficiency parameters, including MFU $e_c$, MBU $e_m$, and communication efficiency $e_+$. These parameters are essential for accurately estimating computation and communication costs. To determine appropriate value for $e_+$, linear regressions are conducted exploit the linear relationship of transmission time \eqref{eq:trans} with the product of $b$, $s$ and $h$. To determine appropriate values for $e_c$ and $e_m$, we aligned the intermediate results of \textit{BestServe}'s simulator, such as Figures \ref{fig:dis-sim-vis} and \ref{fig:col-sim-vis}, with real inference data obtained from manual benchmarking. Through this process, we empirically derived the following values:  
\begin{itemize}[noitemsep]
    \item For the prefill phase: $e_c = 0.65$, $e_m = 0.6$, $e_+ = 0.6$.
    \item For the decode phase: $e_c = 0.65$, $e_m = 0.3$, $e_+ = 0.3$.
\end{itemize}
These values are only roughly tuned to reflect the practical efficiency of the hardware and software environment used in our experiments without losing too much generality.

\subsection{Implementation}
\textit{BestServe} is implemented as a standalone Python module comprising approximately 3,000 lines of code. The implementation is structured around three core components: the \textit{Estimator}, \textit{Simulator}, and \textit{Optimizer}, as detailed in Algorithms \ref{alg:est}-\ref{alg:feasible}. 

The module is designed with modularity and extensibility in mind, allowing each component to be independently improved or replaced. The \textit{Estimator} predicts operator-level latencies using an adapted roofline model, the \textit{Simulator} models the temporal dynamics of request processing, and the \textit{Optimizer} systematically explores serving strategies to identify the configuration that maximizes goodput with adherence to SLO constraints.

To ensure computational efficiency, caching mechanisms are employed in the \textit{Estimator} to avoid redundant calculations, and heuristic approximations are used in the \textit{Simulator} to balance accuracy and performance.

This lightweight design ensures that \textit{BestServe} can run efficiently on standard CPUs, making it accessible for early-stage deployment planning without the need of HPC.

\subsection{Comparison}
\begin{figure*}[!ht]
    \begin{subfigure}[c]{0.495\textwidth}
        \centerline{\includegraphics[width=0.99\textwidth]{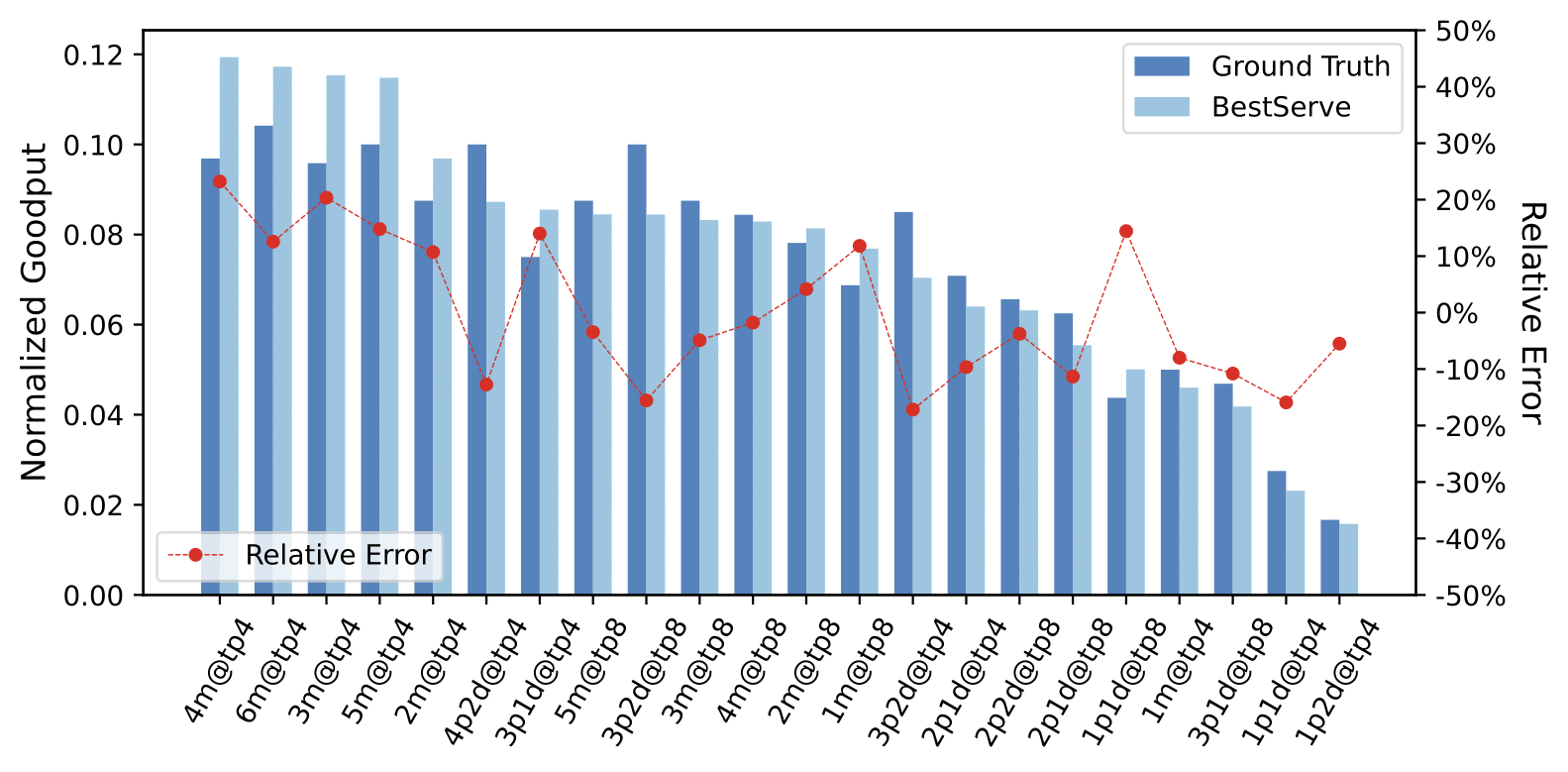}}
        \caption{Operating scenario 1. Average absolute relative error of prediction is $11.2\%$.}
        \label{fig:experiment-op1}
    \end{subfigure}\hfill
    \begin{subfigure}[c]{0.495\textwidth}
        \centerline{\includegraphics[width=0.99\textwidth]{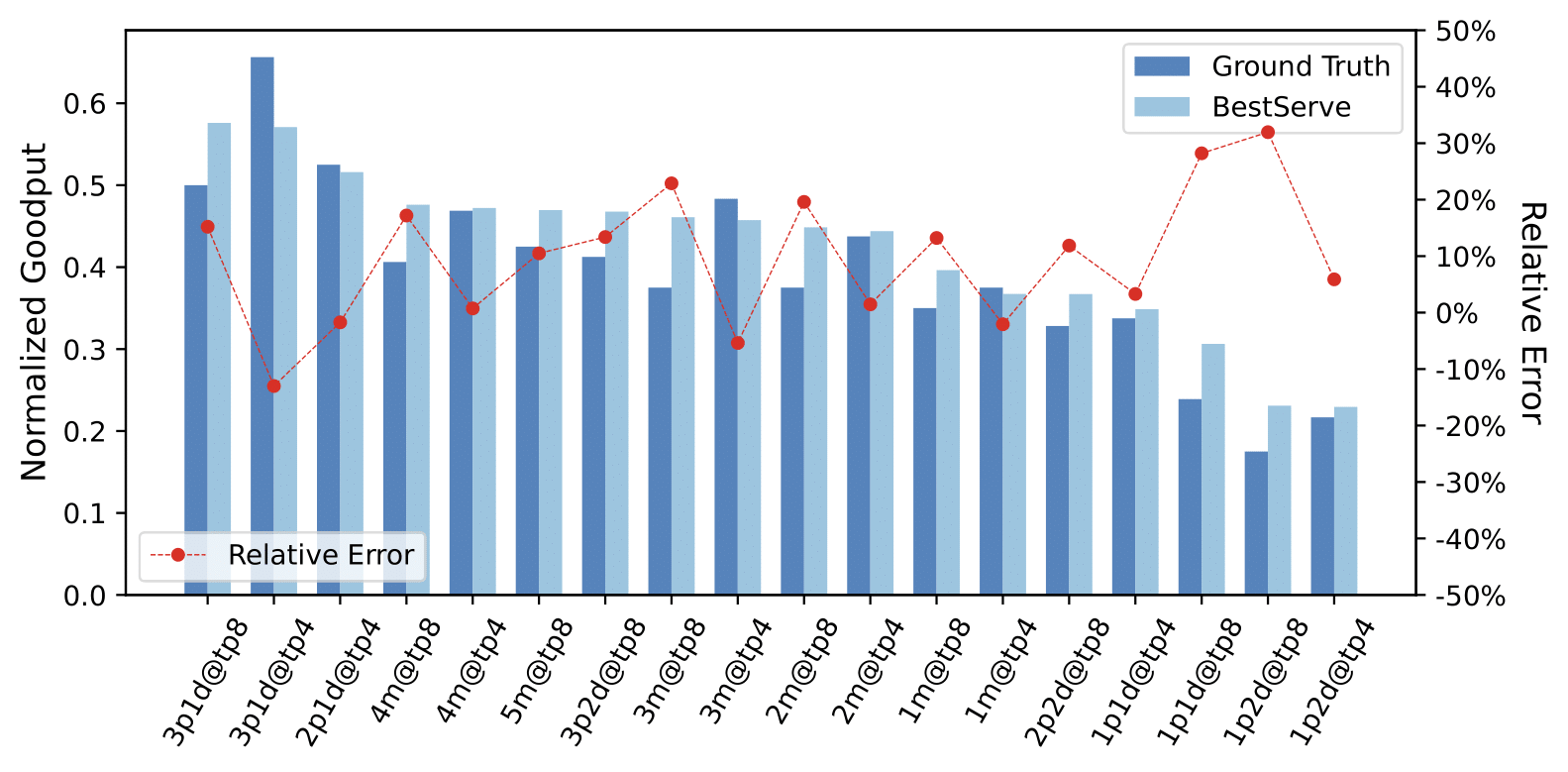}}
        \caption{Operating scenario 2. Average absolute relative error of prediction is $12.1\%$.}
        \label{fig:experiment-op2}
    \end{subfigure}%
    \newline
    \begin{subfigure}[c]{0.495\textwidth}
        \centerline{\includegraphics[width=0.99\textwidth]{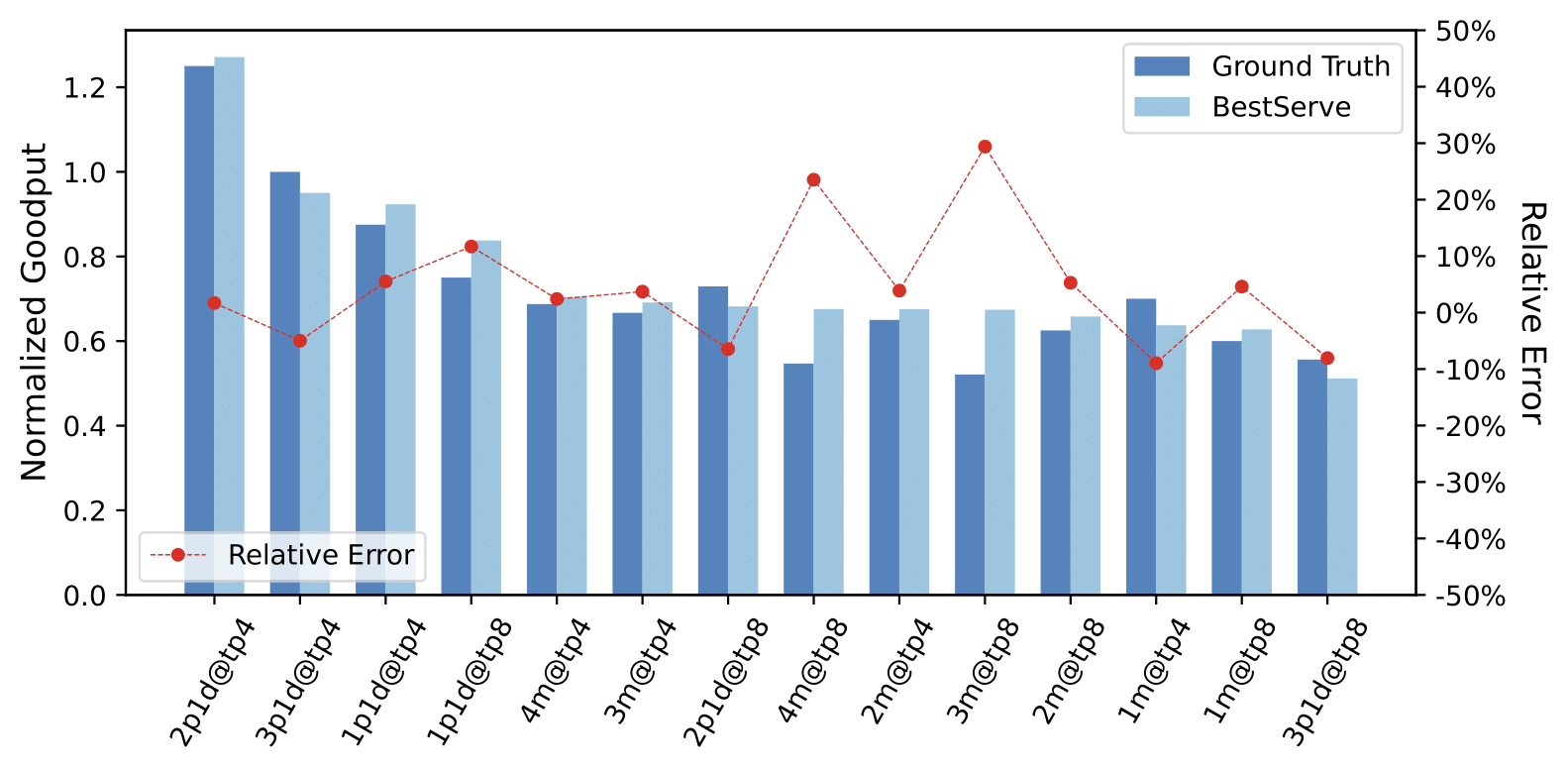}}
        \caption{Operating scenario 3. Average absolute relative error of prediction $8.6\%$.}
        \label{fig:experiment-op3}
    \end{subfigure} \hfill
    \begin{subfigure}[c]{0.495\textwidth}
        \centerline{\includegraphics[width=0.99\textwidth]{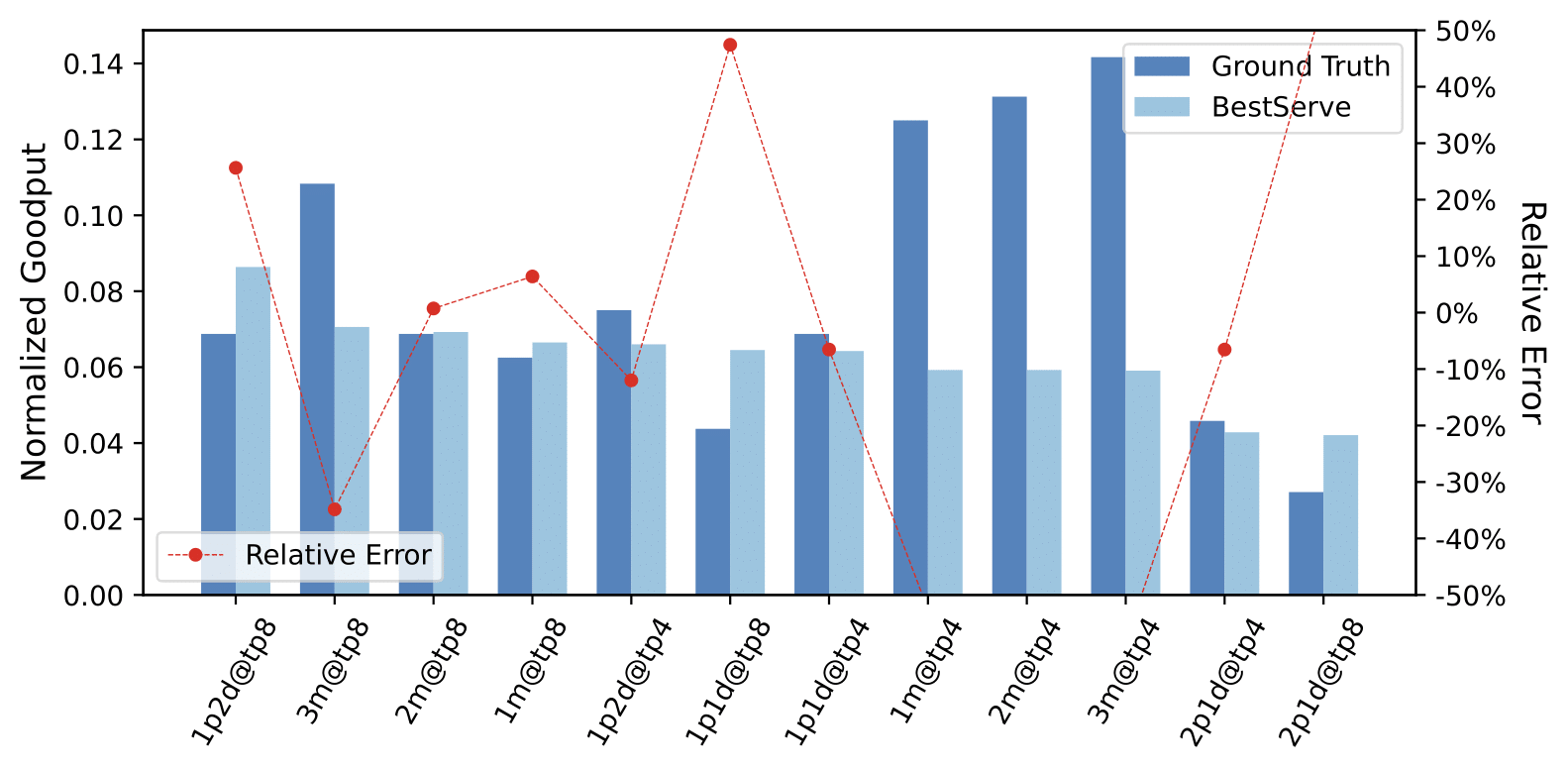}}
        \caption{Operating scenario 4. Average absolute relative error of prediction $30.1\%$.}
        \label{fig:experiment-op4}
    \end{subfigure}

    \caption{Comparison of the maximum goodput estimated by \textit{BestServe} and the ground truth obtained from manual benchmarking. The $x$-axis lists different serving strategies, consisting of instance number and tensor parallel size. The $y$-axis is the normalized goodput, which is the goodput divided by the number of GPUs used in the serving strategy. The red line chart shows the relative error of prediction given by \textit{BestServe}. The histogram is sorted in descending order of the goodput predicted by \textit{BestServe}.}
    \label{fig:experiment}
\end{figure*}

We evaluate how well \textit{BestServe} can estimate the maximum goodput of LLM inference through a series of experiments. Using the ground truth obtained from manual benchmarking, we query \textit{BestServe} to approximate the maximum goodput for each serving strategy across all four operating scenarios. The results are presented as histograms in Figure \ref{fig:experiment}, which compare the normalized goodput predicted by \textit{BestServe} with the ground truth.

The $x$-axis of each histogram lists different serving strategies, characterized by the number of instances and tensor parallel sizes, while the $y$-axis represents the normalized goodput. The red line chart overlays the relative error of \textit{BestServe}'s predictions, providing a clear visualization of its accuracy.

Across the four operating scenarios, the average absolute relative error of \textit{BestServe}'s predictions is 11.2\%, 12.1\%, 8.6\%, and 30.1\% for OP1, OP2, OP3, and OP4, respectively. While the error is moderate in the first three cases, it is notably higher in OP4. This discrepancy is attributed to the simplifications made in the decode phase simulation, such as the use of pseudo batch size, which trades accuracy for computational efficiency. Despite this limitation, \textit{BestServe} demonstrates reasonable accuracy for the majority of scenarios.

% \begin{center}
%     \color{red}
%     [TODO: Elapsed time for generating estimates in Figure 12.]
% \end{center}

It is important to emphasize that \textit{BestServe} achieves these results within mere minutes on a standard CPU, in stark contrast to the extensive time and computational resources required for manual benchmarking on an HPC cluster. This efficiency makes \textit{BestServe} a practical tool for early-stage deployment planning, significantly reducing the burden of trial-and-error benchmarking in the industry.

\section{Limitations} \label{sec:limitations}
\textit{BestServe} is a framework that provides fast and reasonable estimation of the maximum goodput of LLM inference using different strategies. However, it is certainly not without limitations. 

\noindent
\textbf{Over-simplification in decode phase.}\; One of the key simplifications in \textit{BestServe} is the use of the pseudo batch size $b^\dagger$ to approximate the workload of decode instances. While this heuristic significantly reduces the computational cost of simulations, it introduces inaccuracies in scenarios with long generation lengths, such as OP4. In these cases, the decode phase becomes more sensitive to the dynamic nature of continuous batching, and the pseudo batch size heuristic fails to fully capture the complexities of token-level interactions.

This limitation is particularly evident in OP4, where the average absolute relative error of \textit{BestServe}'s predictions reaches 30.1\%. The discrepancy arises because the decode phase simulation assumes a static approximation of batch sizes, which does not account for the variability introduced by long generation lengths. This results in an inaccurate estimation of the workload and, consequently, an overestimation of the maximum goodput, especially for serving strategies using the collocation architecture.

To address this issue, future work could explore more granular simulation techniques that operate at the token level rather than the request level. Although this would increase computational overhead, it could provide a more accurate representation of the decode phase dynamics, particularly in scenarios with long generation lengths.

\noindent
\textbf{Determination of efficiency parameters.}\; The accuracy of \textit{BestServe}'s \textit{Estimator} relies on three key efficiency parameters: MFU $e_c$, MBU $e_m$, and communication efficiency $e_+$. These parameters are determined by the specific hardware and software environment and are critical for accurately estimating computation and communication costs. However, obtaining precise values for these parameters is challenging, as they vary significantly across different systems.

In this work, we use a set of empirically derived default values for these parameters, which provide reasonable estimates for our experimental setup. However, these values are not universally applicable and must be tuned to match the characteristics of the target environment. Proper tuning typically requires minimal profiling and benchmarking of the hardware and software stack, which can be labor-intensive.

This challenge is not unique to \textit{BestServe} but is a common issue in AI infrastructure simulation. Future work could focus on developing automated methods to dynamically determine these parameters based on real-time profiling data. Such advancements would enhance the adaptability and accuracy of \textit{BestServe}, making it more robust across diverse environments.

\noindent
\textbf{Memory insensitivity.}\; \textit{BestServe} is not designed to account for memory consumption, making it memory-insensitive. As a result, certain serving strategies may be deemed feasible by \textit{BestServe} but could fail in practice due to insufficient memory capacity on the underlying hardware. This limitation is particularly relevant when advanced memory management systems, such as PagedAttention in vLLM \cite{10.1145/3600006.3613165}, are employed. These systems dynamically manage memory to optimize performance, introducing complexities that \textit{BestServe} does not currently model. Addressing memory constraints and incorporating memory-aware simulations is also a promising direction in the future.

\section{Conclusion and Perspectives} \label{sec:conclusion}

We propose a novel framework, \textit{BestServe}, for determining the optimal LLM serving strategy to maximize goodput. The framework is hierarchically structured with three layers: an \textit{Estimator}, which predicts the time costs of a single LLM forward pass using an adapted roofline model; a \textit{Simulator}, which models the latency of processing asynchronously arriving requests; and an \textit{Optimizer}, which systematically searches for the optimal serving strategy that achieves the highest goodput while adhering to SLO constraints. Compared to conventional trial-and-error approaches, \textit{BestServe} is significantly more efficient and requires minimal computational resources. We validate the framework by comparing its predictions with real inference tests conducted under various serving strategies and operating scenarios, demonstrating its effectiveness in providing moderately accurate and fast results.

Our work opens up several promising directions for future research. One critical area is the refinement of the decode phase simulation. The current reliance on the pseudo batch size heuristic, while computationally efficient, introduces inaccuracies in scenarios with long generation lengths. Developing more sophisticated methods, such as token-level simulations or other advanced techniques, could significantly enhance the simulator's accuracy and better capture the complexities of continuous batching.

Another important direction is the generalization of \textit{BestServe} to broader contexts. Extending the framework to support more advanced LLM architectures and serving techniques, such as mixture of experts (MoE) \cite{jiang2024mixtralexperts} and pipeline parallelism (PP) \cite{10.5555/3454287.3454297,harlap2018pipedreamfastefficientpipeline}, would greatly expand its applicability. Furthermore, adapting \textit{BestServe} to accommodate diverse hardware configurations and other model families would make it a more versatile and widely applicable tool for the AI community.

From a theoretical perspective, there is considerable potential to deepen the understanding of LLM inference by modeling it through more sophisticated queueing systems. While LLM inference naturally aligns with queueing theory, simplified models such as $M/D/c$ fail to capture the intricacies of real-world scenarios. Developing advanced queueing models and analyzing their theoretical foundations could provide valuable insights into the dynamics of LLM inference and improve the predictive power of the framework. Similarly, the selection of the relaxation factor $\tau$ in request rate feasibility checks remains an open challenge. This parameter, which balances accuracy and robustness, could be analyzed using statistical methods. Addressing these challenges would further enhance the adaptability and reliability of \textit{BestServe}, paving the way for more robust and efficient LLM serving strategies.

The impact of \textit{BestServe} extends beyond its immediate application to LLM serving. By significantly reducing the computational and time costs associated with benchmarking, \textit{BestServe} enables service providers to rapidly explore and deploy optimal serving strategies, even in resource-constrained environments. Its lightweight design and modular extensibility make it a practical tool for early-stage deployment planning, helping organizations efficiently meet user demands while adhering to SLOs. Furthermore, \textit{BestServe} lays a strong foundation for future innovations in AI infrastructure simulation, offering a scalable framework that can adapt to emerging LLM architectures and advanced serving techniques. As the field of LLMs continues to evolve, \textit{BestServe} has the potential to play a pivotal role in shaping the next generation of efficient and reliable AI serving systems.

\section{Related Works}

\textbf{PD Disaggregation.}\; Chronologically, Splitwise \cite{patel2024splitwiseefficientgenerativellm} was the first to propose the idea of disaggregating prefill and decode phases. TetriInfer \cite{hu2024inferenceinterferencedisaggregatellm} explored an effective scheduling algorithm to prevent decode requests from clustering. Following up, this field saw DistServe \cite{Zhong2024DistServeDP} introduce a heuristic algorithm to automatically determine the parallelism configuration, while Mooncake \cite{qin2024mooncakekvcachecentricdisaggregatedarchitecture} became the first holistic and refined disaggregation serving architecture. Whether disaggregation outperforms collocation is really the reason why we design \textit{BestServe} to consider both collocation and disaggregation architectures.

\noindent
\textbf{Queueing Theory Perspective of LLM Inference.}\; To the best of our knowledge, DistServe \cite{Zhong2024DistServeDP} was the first to draw an analogy between the behavior of the prefill phase and an $M/D/1$ queue. While this approach provides valuable insights, it relies on simplified queueing models that fail to capture the complexities of real-world LLM inference. \textit{BestServe} extends this perspective by incorporating a queueing theory-inspired framework with numerical simulations, enabling more accurate modeling of LLM inference dynamics.

\noindent
\textbf{AI Infrastructure Simulation.}\; SimAI \cite{SimAI} is a monumental work on AI infrastructure simulation, focusing on optimizing LLM training pipelines. It models the interactions between hardware and software components to identify bottlenecks and improve training performance. While SimAI emphasizes training, \textit{BestServe} complements it by addressing the unique challenges of LLM inference. Both frameworks share a simulation-based approach and a hardware-aware design, but \textit{BestServe} specifically targets goodput optimization for serving strategies, making it a practical tool for early-stage deployment planning.

%-------------------------------------------------------------------------------
\section*{Acknowledgments}
This work is based on a project by Xiannan Hu and Tianyou Zeng during their internships at Huawei Cloud, Huawei Technologies Co., Ltd. The authors sincerely thank their on-site and academic supervisors for their valuable support and guidance.
%-------------------------------------------------------------------------------

%-------------------------------------------------------------------------------
\section*{Availability}
Due to Huawei's strict policies, the circulation of code and data is prohibited. Consequently, we are unable to provide additional information beyond what is presented in this paper. However, we hope that the detailed explanations and pseudocode included herein will serve as a valuable resource for understanding and implementing the concepts discussed.
%-------------------------------------------------------------------------------

\appendix
\section{Work and Memory Traffics of Modules in LLMs from the LLaMa Family}
\label{appendix:est}
In this section, we list out the amount of work and memory traffics that we have estimated for each modules of LLaMa in both phases. We reserved the following symbols for the following model parameters:
\begin{itemize}[noitemsep]
    \item $b$: batch size,
    \item $s$: sequence length,
    \item $h$: hidden size of dimension,
    \item $h_0$: intermediate size of MLP,
    \item $h_{kv}$: number of key-value heads,
    \item $h_q$: number of query heads.
\end{itemize}
We also reserve the following symbols for hardware specification and efficiency parameters:
\begin{itemize}[noitemsep]
    \item $S_c$: peak flop,
    \item $S_m$: peak memory bandwidth,
    \item $S_+$: peak communication bandwidth,
    \item $e_c$: MFU,
    \item $e_m$: MBU,
    \item $e_+$: communication efficiency.
\end{itemize}

\subsection{Normalization}
LLaMa adopts root mean square normalization (RMSNorm) \cite{zhang2019rootmeansquarelayer}.
\begin{itemize}
    \item For prefill phase, we count the amount of works and memory traffics as in Table \ref{tab:rms-est-p}.

    \begin{table}[!ht]
        \centering
        \small
        \begin{tabular}{| c | c | c | c |} 
        \hline
        $i$ & \textbf{Description}  & $W_i$ (unit: FLOP) & $Q_i$ (unit: byte) \\ 
        \hline\hline
        1 & \texttt{POW} & $bsh$ & $4bsh$ \\ \hline
        2 & \texttt{MEAN} & $bsh$ & $2bsh+2bs$ \\ \hline
        3 & \texttt{ADD} & $bs$ & $4bs$ \\ \hline
        4 & \texttt{RSQRT} & $bs$ & $4bs$ \\ \hline
        5 & \texttt{MUL} & $bsh$ & $4bsh+2bs$ \\ \hline
        6 & \texttt{MUL} & $bsh$ & $4bsh+2h$ \\ \hline
        \end{tabular}
        \caption{Works and memory traffics of each operation in the prefill phase of a LLaMa RMSNorm module.}
        \label{tab:rms-est-p}
    \end{table}

    We use the following formula to estimate the computation time of the prefill phase of RMSNorm module:
    \begin{equation} \label{eq:rms-est}
        T_{\text{RMSNorm}} = \frac{W_1}{P_1} + \frac{W_2}{P_2} + \frac{W_3}{P_3} + \frac{W_4}{P_4} + \frac{W_5}{P_5} + \frac{W_6}{P_6},
    \end{equation}
    where $P_i := \min\{I^*, I_i\}e_mS_m$, $I_i := W_i/Q_i$ for all $i = 1,2,3,4,5,6$.

    \item For decode phase, we modify the amount of works and memory traffics as in Table \ref{tab:rms-est-d}.
    \begin{table}[!ht]
        \centering
        \small
        \begin{tabular}{| c | c | c | c |} 
        \hline
        $i$ & \textbf{Description}  & $W_i$ (unit: FLOP) & $Q_i$ (unit: byte) \\ 
        \hline\hline
        1 & \texttt{POW} & $bh$ & $4bh$ \\ \hline
        2 & \texttt{MEAN} & $bh$ & $2bh+2b$ \\ \hline
        3 & \texttt{ADD} & $b$ & $4b$ \\ \hline
        4 & \texttt{RSQRT} & $b$ & $4b$ \\ \hline
        5 & \texttt{MUL} & $bh$ & $4bh+2b$ \\ \hline
        6 & \texttt{MUL} & $bh$ & $4bh+2h$ \\ \hline
        \end{tabular}
        \caption{Works and memory traffics of each operation in the decode phase of a LLaMa RMSNorm module.}
        \label{tab:rms-est-d}
    \end{table}

    We use the same formula \eqref{eq:rms-est} as in the prefill phase to estimate the computation time of the prefill phase of RMSNorm module.
\end{itemize}

\subsection{Attention}
The normal Attention module in LLaMa follows Lines 242-290 of \cite{modellingllama}.

\begin{itemize}
    \item For prefill phase, we count the amount of works and memory traffics as in Table \ref{tab:attention-est}.
    \begin{table*}[!ht]
        \centering
        \small
        \begin{tabular}{| c | c | c | c |} 
        \hline
        $i$ & \textbf{Description}  & $W_i$ (unit: FLOP) & $Q_i$ (unit: byte) \\ 
        \hline\hline
        1 & \texttt{Q\_PROJ} & $2bsh^2$ & $2(2bsh+h^2)$ \\ \hline
        2 & \texttt{K\_PROJ} & $2bsh^2h_{kv}/h_{q}$ & $2(bsh+h^2h_{kv}/h_{q}+bshh_{kv}/h_{q})$ \\ \hline
        3 & \texttt{V\_PROJ} & $2bsh^2h_{kv}/h_{q}$ & $2(bsh+h^2h_{kv}/h_{q}+bshh_{kv}/h_{q})$ \\ \hline
        4 & \texttt{RoPE} & $3.5bsh(1+h_{kv}/h_{q})$ & $2bsh(8.5+8.5h_{kv}/h_{q}+2/h_{q})$ \\ \hline
        5 & $\texttt{QK}^\top$ & $2bs^2h$ & $2(2bsh+bh_{q}s^2)$ \\ \hline
        6 & \texttt{div} & $bh_{q}s^2$ & $4bh_{q}s^2$ \\ \hline
        7 & \texttt{add} & $bh_{q}s^2$ & $2(2bh_{q}s^2 +bs^2)$ \\ \hline
        8 & \texttt{softmax} & $3bh_{q}s^2$ & $4bh_{q}s^2$ \\ \hline
        9 & \texttt{@\,V} & $2bs^2h$ & $2(bh_{q}s^2 + 2bsh)$ \\ \hline
        10 & \texttt{O\_PROJ} & $2bsh^2$ & $2(2bsh+h^2)$ \\ \hline
        \end{tabular} 
        \caption{Works and memory traffics of each operation in the prefill phase of a LLaMa Attention module.}
        \label{tab:attention-est}
    \end{table*}
    \begin{table*}[!ht]
        \centering
        \small
        \begin{tabular}{| c | c | c | c |} 
        \hline
        $i$ & \textbf{Description}  & $W_i$ (unit: FLOP) & $Q_i$ (unit: byte) \\ 
        \hline\hline
        1 & \texttt{Q\_PROJ} & $2bh^2$ & $2(2bh+h^2)$ \\ \hline
        2 & \texttt{K\_PROJ} & $2bh^2h_{kv}/h_{q}$ & $2(bh+h^2h_{kv}/h_{q}+bhh_{kv}/h_{q})$ \\ \hline
        3 & \texttt{V\_PROJ} & $2bh^2h_{kv}/h_{q}$ & $2(bh+h^2h_{kv}/h_{q}+bhh_{kv}/h_{q})$ \\ \hline
        4 & \texttt{RoPE} & $3.5bh(1+h_{kv}/h_{q})$ & $2bh(8.5+8.5h_{kv}/h_{q}+2/h_{q})$ \\ \hline
        5 & \texttt{update} & - & $4bshh_{kv}/h_{q}$ \\ \hline
        $\star$ & \texttt{repeat\_kv} & - & $4bsh(1+h_{kv}/h_{q})$ \\ \hline
        6 & $\texttt{QK}^\top$ & $2bsh$ & $2b(h+hs+h_{q}s)$ \\ \hline
        7 & \texttt{div} & $bh_{q}s$ & $4bh_{q}s$ \\ \hline
        8 & \texttt{add} & $bh_{q}s$ & $2(2bh_{q}s +bs)$ \\ \hline
        9 & \texttt{upcast} & - & $4bh_{q}s$ \\ \hline
        10 & \texttt{softmax} & $3bh_{q}s$ & $4bh_{q}s$ \\ \hline
        11 & \texttt{@\,V} & $2bsh$ & $2b(h+hs+h_{q}s)$ \\ \hline
        12 & \texttt{O\_PROJ} & $2bh^2$ & $2(2bh+h^2)$ \\ \hline
        \end{tabular}
        \caption{Works and memory traffics of each operation in the decode phase of a LLaMa Attention module.}
        \label{tab:attention-est-decode}
    \end{table*}

    We use the following formula to estimate the computation time of the prefill phase of Attention module:
    \begin{multline} \label{eq:attention-est}
        T_\textrm{Attention} = \frac{W_1}{P_1} + \frac{W_2}{P_2} + \frac{W_3}{P_3} + \frac{W_4}{P_4} + \frac{W_5}{P_5} + \\ \frac{W_6}{P_6} + \frac{W_7}{P_7} + \frac{W_8}{P_8} + \frac{W_9}{P_9} + \frac{W_{10}}{P_{10}},
    \end{multline}
    where $P_i := \min\{I_i, I^*\} e_mS_m$, $I_i := W_i/Q_i$ for all $i = 1,2,3,4,5,6,7,8,9,10$.
    \item For the decode phase, the presence of KV-Cache greatly reduces the computation effort, therefore several operations that are not of computation nature becomes dominated. Among the many of them, the following three seems  to be representative: the update of KV-Cache, repetition of KV heads, and upcast of tensors, arriving at Table \ref{tab:attention-est-decode}.

    \begin{align} \label{eq:attention-est-decode}
    \begin{aligned}
        T_\mathrm{Attention}^+ = & \frac{W_1}{P_1} + \frac{W_2}{P_2} + \frac{W_3}{P_3} + \frac{W_4}{P_4} + \frac{Q_5}{\kappa_\mathrm{update}} \\ 
        & + \texttt{Is\_GQA} \times \frac{Q_\star}{\kappa_{\mathrm{kv}}} + \frac{W_6}{P_6} + \frac{W_7}{P_7} + \frac{W_8}{P_8} \\
        & + \frac{Q_9}{\kappa_\mathrm{upcast}} + \frac{W_{10}}{P_{10}} + \frac{W_{11}}{P_{11}} + \frac{W_{12}}{P_{12}},
    \end{aligned}
    \end{align}
    where $P_i := \min\{I^*, I_i\}e_mS_m$ and $I_i := W_i/Q_i$ for all $i=1,2,3,4,6,7,8,10,11,12$. The three hyperparameters $\kappa_\mathrm{update}$, $\kappa_\mathrm{kv}$, $\kappa_\mathrm{upcast}$ are introduced to reflect the rate of updating KV-Cache, repeating KV heads and upcasting floats to \texttt{FP32} format, and \texttt{Is\_GQA} is a boolean coefficient indicating whether the concerned model adopts GQA in its attention module.
\end{itemize}

\subsection{MLP}
LLaMa uses SiLU \cite{elfwing2017sigmoidweightedlinearunitsneural} as the activation function for its MLP.
\begin{itemize}
    \item For prefill phase, we count the amount of works and memory traffics as in Table \ref{tab:mlp-est}. We use \eqref{eq:mlp-est} to estimate the computation time of the prefill phase of MLP module.

    \item For decode phase, we modify the amount of works and memory traffics as in Table \ref{tab:mlp-est-decode}. The same formula \eqref{eq:mlp-est} applies to the decode phase of MLP module.
\end{itemize}

\section{Work and Memory Traffics of Modules in LLMs from the LLaMa Family When TP is Considered}
\label{appendix:est-tp}

If the number of tensor replicas involved in the computation is $t$, we can accordingly adjust works and memory traffics of each module.

\begin{table*}[!ht]
    \centering
    \small
    \begin{tabular}{| c | c | c | c |} 
    \hline
    $i$ & \textbf{Description}  & $W_i$ (unit: FLOP) & $Q_i$ (unit: byte) \\ 
    \hline\hline
    1 & \texttt{Q\_PROJ} & $2bsh^2/t$ & $2(2bsh+h^2)/t$ \\ \hline
    2 & \texttt{K\_PROJ} & $2bsh^2h_{kv}/(th_{q})$ & $2(bsh+h^2h_{kv}/(th_{q})+bshh_{kv}/(th_{q}))$ \\ \hline
    3 & \texttt{V\_PROJ} & $2bsh^2h_{kv}/(th_{q})$ & $2(bsh+h^2h_{kv}/(th_{q})+bshh_{kv}/(th_{q}))$ \\ \hline
    4 & \texttt{RoPE} & $3.5bsh(1+h_{kv}/h_{q})$ & $2bsh(8.5+8.5h_{kv}/h_{q}+2/h_{q})$  \\ \hline
    5 & ${QK}^T$ & $2bs^2h / t$ & $2(2bsh+bh_{q}s^2)/t$ \\ \hline
    6 & \texttt{div} & $bh_{q}s^2 / t$ & $4bh_{q}s^2/t$ \\ \hline
    7 & \texttt{add} & $bh_{q}s^2 / t$ & $2(2bh_{q}s^2/t +bs^2)$ \\ \hline
    8 & \texttt{softmax} & $3bh_{q}s^2 / t$ & $4bh_{q}s^2/t$ \\ \hline
    9 & \texttt{@\,V} & $2bs^2h / t$ & $2(bh_{q}s^2 + 2bsh)/t$ \\ \hline
    10 & \texttt{O\_PROJ} & $2bsh^2 / t$ & $2(bsh+bsh/t+h^2)$ \\ \hline
    \end{tabular}
    \caption{Works and memory traffics of each operation in the prefill phase of a LLaMa Attention module considering TP.}
    \label{tab:attention-est-tp}
\end{table*}
\begin{table*}[!ht]
    \centering
    \small
    \begin{tabular}{| c | c | c | c |} 
    \hline
    $i$ & \textbf{Description}  & $W_i$ (unit: FLOP) & $Q_i$ (unit: byte) \\ 
    \hline\hline
    1 & \texttt{Q\_PROJ} & $2bh^2/t$ & $2(2bh+h^2)/t$ \\ \hline
    2 & \texttt{K\_PROJ} & $2bh^2h_{kv}/h_{q}$ & $2(bh+h^2h_{kv}/(th_{q})+bhh_{kv}/(th_{q}))$ \\ \hline
    3 & \texttt{V\_PROJ} & $2bh^2h_{kv}/(th_{q})$ & $2(bh+h^2h_{kv}/(th_{q})+bhh_{kv}/(th_{q}))$ \\ \hline
    4 & \texttt{RoPE} & $3.5bh(1+h_{kv}/h_{q})$ & $2bh(8.5+8.5h_{kv}/h_{q}+2/h_{q})$ \\ \hline
    5 & \texttt{update} & - & $2bshh_{kv}/h_{q}/t$ \\ \hline
    $\star$ & \texttt{repeat\_kv} & - & $2bsh(1+h_{kv}/h_{q})/t$ \\ \hline
    6 & $\texttt{QK}^\top$ & $2bsh/t$ & $2b(h+hs+h_{q}s)/t$ \\ \hline
    7 & \texttt{div} & $bh_{q}s/t$ & $4bh_{q}s/t$ \\ \hline
    8 & \texttt{add} & $bh_{q}s/t$ & $2(2bh_{q}s/t +bs)$ \\ \hline
    9 & \texttt{upcast} & - & $4bh_{q}s/t$ \\ \hline
    10 & \texttt{softmax} & $3bh_{q}s$ & $4bh_{q}s/t$ \\ \hline
    11 & \texttt{@\,V} & $2bsh/t$ & $2b(h+hs+h_{q}s)/t$ \\ \hline
    12 & \texttt{O\_PROJ} & $2bh^2/t$ & $2(bh+h^2/t+bh/t)$ \\ \hline
    \end{tabular}
    \caption{Works and memory traffics of each operation in the decode phase of a LLaMa Attention module considering TP.}
    \label{tab:attention-est-decode-tp}
\end{table*}

\subsection{Normalization}
\begin{itemize}
    \item For prefill phase, we still count the amount of works and memory traffics as in Table \ref{tab:rms-est-p}. We also use \eqref{eq:rms-est} to estimate the computation time of the prefill phase of MLP module when TP is considered.

    \item For decode phase, we still count the amount of works and memory traffics as in Table \ref{tab:rms-est-d}. We also use \eqref{eq:rms-est} to estimate the computation time of the decode phase of MLP module when TP is considered.
\end{itemize}

\subsection{Attention}
\begin{itemize}
    \item For prefill phase, we count the amount of works and memory traffics as in Table \ref{tab:attention-est-tp}.

    We still use \eqref{eq:attention-est} to estimate the computation time of the prefill phase of Attention module when TP is considered.

    \item For the decode phase, we count the amount of works and memory traffics as in Table \ref{tab:attention-est-decode-tp}.

    We still use \eqref{eq:attention-est-decode} to estimate the computation time of the prefill phase of Attention module when TP is considered.
\end{itemize}

\subsection{MLP}

\begin{table}[!ht]
    \centering
    \small
    \begin{tabular}{| c | c | c | c |} 
     \hline
     $i$ & \textbf{Description}  & $W_i$ (unit: FLOP) & $Q_i$ (unit: byte) \\ 
     \hline\hline
     1 & \texttt{GATE\_PROJ} & $2bshh_0/t$ & $2(bs(h+h_0) + hh_0)/t$ \\ \hline
     2 & \texttt{SiLU} & $5bsh_0/t$ & $4bsh_0/t$ \\ \hline
     3 & \texttt{UP\_PROJ} & $2bshh_0/t$ & $2(bs(h+h_0) + hh_0)/t$ \\ \hline
     4 & \texttt{mul} & $bsh_0/t$ & $6bsh_0/t$ \\ \hline
     5 & \texttt{DOWN\_PROJ} & $2bshh_0/t$ & $2(bs(h+h_0) + hh_0)/t$ \\ \hline
     6 & \texttt{add} & $bsh/t$ & $4bsh_0/t$ \\ \hline
    \end{tabular}
    \caption{Works and memory traffics of each operation in the prefill phase of a LLaMa MLP module considering TP.}
    \label{tab:mlp-est-tp}
\end{table}
\begin{table}[!ht]
    \centering
    \small
    \begin{tabular}{| c | c | c | c |} 
     \hline
     $i$ & \textbf{Description}  & $W_i$ (unit: FLOP) & $Q_i$ (unit: byte) \\ 
     \hline\hline
     1 & \texttt{GATE\_PROJ} & $2bhh_0/t$ & $2(b(h+h_0) + hh_0)/t$ \\ \hline
     2 & \texttt{SiLU} & $5bh_0/t$ & $4bh_0/t$ \\ \hline
     3 & \texttt{UP\_PROJ} & $2bhh_0/t$ & $2(b(h+h_0) + hh_0)/t$ \\ \hline
     4 & \texttt{mul} & $bh_0/t$ & $6bsh_0/t$ \\ \hline
     5 & \texttt{DOWN\_PROJ} & $2bhh_0/t$ & $2(b(h+h_0) + hh_0)/t$ \\ \hline
     6 & \texttt{add} & $bh/t$ & $4bh_0/t$ \\ \hline
    \end{tabular}
    \caption{Works and memory traffics of each operation in the decode phase of a LLaMa MLP module considering TP.}
    \label{tab:mlp-est-decode-tp}
\end{table}

\begin{itemize}
    \item For prefill phase, we count the amount of works and memory traffics as in Table \ref{tab:mlp-est-tp}. We still use \eqref{eq:mlp-est} to estimate the computation time of the prefill phase of MLP module.
    
    \item For decode phase, we modify the amount of works and memory traffics as in Table \ref{tab:mlp-est-decode-tp}. The same formula \eqref{eq:mlp-est} applies to the decode phase of MLP module.

\end{itemize}

%-------------------------------------------------------------------------------
\bibliographystyle{plain}
\raggedright
\bibliography{references}

%%%%%%%%%%%%%%%%%%%%%%%%%%%%%%%%%%%%%%%%%%%%%%%%%%%%%%%%%%%%%%%%%%%%%%%%%%%%%%%%
\end{document}